\documentclass[10pt,journal,compsoc]{IEEEtran}
\usepackage{amsmath,amssymb,amsfonts}%
\usepackage{algorithmic}
\usepackage{graphicx}
\usepackage{textcomp}
\usepackage{url}
\usepackage{color}
\usepackage{balance}
\usepackage{booktabs}
\usepackage[table]{xcolor} 
\usepackage{rotating}
\usepackage{amsfonts}
\usepackage{graphicx}
\usepackage{multirow}
\usepackage{multicol}
\usepackage{float}
\usepackage{mathtools}
\DeclarePairedDelimiter{\norm}{\lVert}{\rVert}
\usepackage[thinlines]{easytable}
\usepackage{graphicx,subcaption}

\usepackage{graphicx}

\ifCLASSOPTIONcompsoc
  \usepackage[nocompress]{cite}
\else
  \usepackage{cite}
\fi

\ifCLASSINFOpdf

\else

\fi

\begin{document}

\title{Video Description: A Survey of Methods, Datasets and Evaluation Metrics}

\author{Nayyer~Aafaq,
        Ajmal~Mian,
        Wei~Liu,
        Syed~Zulqarnain~Gilani,
        and~Mubarak~Shah

\IEEEcompsocitemizethanks{\IEEEcompsocthanksitem N. Aafaq, S. Z. Gilani, W. Liu and A. Mian are with the Department of Computer Science and Software Engineering (CSSE), University of Western Australia (UWA), WA,
6009.\protect
\IEEEcompsocthanksitem M. Shah is with University of Central Florida (UCF), FL, USA\protect \\

E-mail: nayyer.aafaq@research.uwa.edu.au, {[zulqarnain.gilani, wei.liu, ajmal.mian]}@uwa.edu.au, shah@crcv.ucf.edu
}
}

\IEEEtitleabstractindextext{%
\begin{abstract}
Video description is the automatic generation of natural language sentences that describe the contents of a given video. It has applications in human-robot interaction, helping the visually impaired and video subtitling. The past few years have seen a surge of research in this area due to the unprecedented success of deep learning in computer vision and natural language processing. Numerous methods, datasets and evaluation metrics have been proposed in the literature, calling the need for a comprehensive survey to focus research efforts in this flourishing new direction. This paper fills the gap by surveying the state of the art approaches with a focus on deep learning models; comparing benchmark datasets in terms of their domains, number of classes, and repository size; and identifying the pros and cons of various evaluation metrics like  SPICE, CIDEr, ROUGE, BLEU, METEOR, and WMD. Classical video description approaches combined subject, object and verb detection with template based language models to generate sentences. However, the release of large datasets revealed that these methods can not cope with the diversity in unconstrained open domain videos. Classical approaches were followed by a very short era of statistical methods which were soon replaced with deep learning, the current state of the art in video description. Our survey shows that despite the fast-paced developments, video description research is still in its infancy due to the following reasons. Analysis of video description models is challenging because it is difficult to ascertain the contributions, towards accuracy or errors, of the visual features and the adopted language model in the final description. Existing datasets neither contain adequate visual diversity nor complexity of linguistic structures. Finally, current evaluation metrics fall short of measuring the agreement between machine generated descriptions with that of humans.  We conclude our survey by listing promising future research directions.
\end{abstract}

\begin{IEEEkeywords}
Deep learning, video description, video captioning, video to text, language in vision, video captioning datasets, video captioning evaluation metrics, BLEU, METEOR, ROUGE, CIDEr, SPICE, WMD.
\end{IEEEkeywords}}

\maketitle

\IEEEdisplaynontitleabstractindextext

\IEEEpeerreviewmaketitle

\ifCLASSOPTIONcompsoc
\IEEEraisesectionheading{\section{Introduction}\label{sec:introduction}}
\else
\section{Introduction}
\label{sec:introduction}
\fi

\IEEEPARstart{D}{escribing} a short video in natural language is a trivial task for most people, but a very challenging one for machines. Automatic video description involves understanding of many entities and the detection of their occurrences in a video employing computer vision techniques. These entities include \textit{background scene, humans, objects, human actions, human-object interactions, human-human interactions, other events}, and the \textit{order} in which events occur. All this information must then be articulated using a comprehensible and grammatically correct text employing Natural Language Processing (NLP) techniques. Over the past few years, these two traditionally independent fields, Computer Vision (CV) and Natural Language Processing (NLP) have joined forces to address the upsurge of research interests in understanding and describing images and videos. Special issues of journals are published focusing on language in vision~\cite{scidirlanginvision} and workshops uniting the two areas have also been held regularly at both NLP and CV conferences~\cite{languageandvisioncvpr2015, languageandvisioncvpr2018, languageandvisioniccv2015, languageandvisionnaacl2018}.


Automatic video description has many applications in human-robot interaction, automatic video subtitling and video surveillance. It can be used to help the visually impaired by generating verbal descriptions of surroundings through speech synthesis, or automatically generating and reading out film descriptions. Currently, these are achieved through very costly and time-consuming manual processes. Another application is the description of sign language videos in natural language. Video description can also generate written procedures for human or service robots by automatically converting actions in a demonstration video into simple instructions, for example, assembling furniture, installing CD-ROM, making coffee or changing a flat tyre~\cite{alayrac2016unsupervised, brand1997inverse}. 

\begin{figure*}[htbp] 
     \centering
     \includegraphics[width=0.8\textwidth]{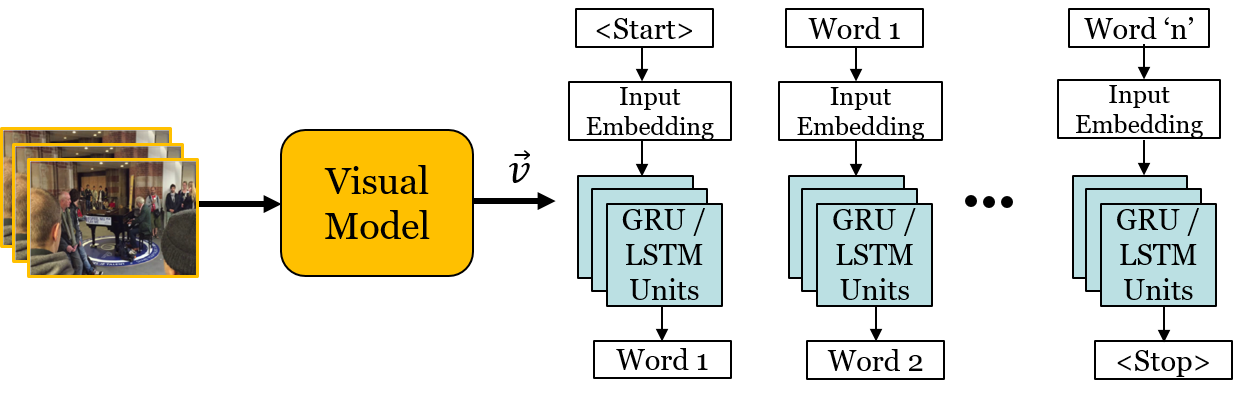}
    \vspace{-3mm}
     \caption{A basic framework for deep learning based video captioning. A visual model encodes the video frames into a vector space. The language model takes input of visual vector and word embeddings to generate the sentence that describes the input visual content.}
\label{fig:basicblockdiagram}
\vspace{-5mm}
\end{figure*}

The advancement of video description opens up enormous opportunities in many application domains. It is envisaged that in the near future, we would be able to interact with robots in the same manner as with humans~\cite{rohrbach2013translating}. If video description is advanced to the stage of being able to comprehend events unfolding in the real world and render them in spoken words, \textit{Service Robots} or \textit{Smart phone Apps} will be able to understand human actions and other events to converse with humans in a much more meaningful and coherent manner. For example, they could answer a user's question as to where they left their wallet or discuss what they should cook for dinner. In industry settings, they could potentially remind a worker of any actions/procedures that are missing from a routine operation. The recent release of a dialogue dataset, \emph{Talk the Walk}~\cite{talkthewalk}, has introduced yet another interesting application where a natural language dialogue between a \emph{guide} and a \emph{tourist} helps the tourist to reach a previously unseen location on a map using perception, action and interaction modeling.

Leveraging the recent developments in deep neural networks for NLP and CV, and the increased availability of large multi-modal datasets, automatically generating stories from pixels is no longer a science fiction. This growing body of work has mainly originated from the robotics community and can be labeled broadly as \textit{language grounded meaning from vision to robotic perception}~\cite{Roy2005b}. Related research areas include, connecting words to pictures \cite{barnard2003matching,berg2004names,deng2009construction}, narrating images in natural language sentences \cite{farhadi2010every,kulkarni2011baby,li2011composing} and understanding natural language instructions for robotic applications \cite{guadarrama2013grounding,matuszek2010following,tellex2011understanding}. 
Another closely related field is \emph{Visual Information Retrieval (VIR)}, which takes visual (image, drawing or sketch), text (tags, keywords or complete sentence) or mixed visual and text query to perform content based search. Thanks to the release of benchmark datasets MS COCO~\cite{lin2014microsoft} and Flicker30k \cite{young2014image}, research in \textit{image captioning and retrieval} \cite{donahue2015long,kiros2014unifying,fang2015captions,mao2015learning}, and \textit{image question answering}  \cite{malinowski2014multi,antol2015vqa,ren2015exploring,yu2015visual} has also become very active.

\begin{figure*}[t] 
     \centering
     \includegraphics[width=\textwidth]{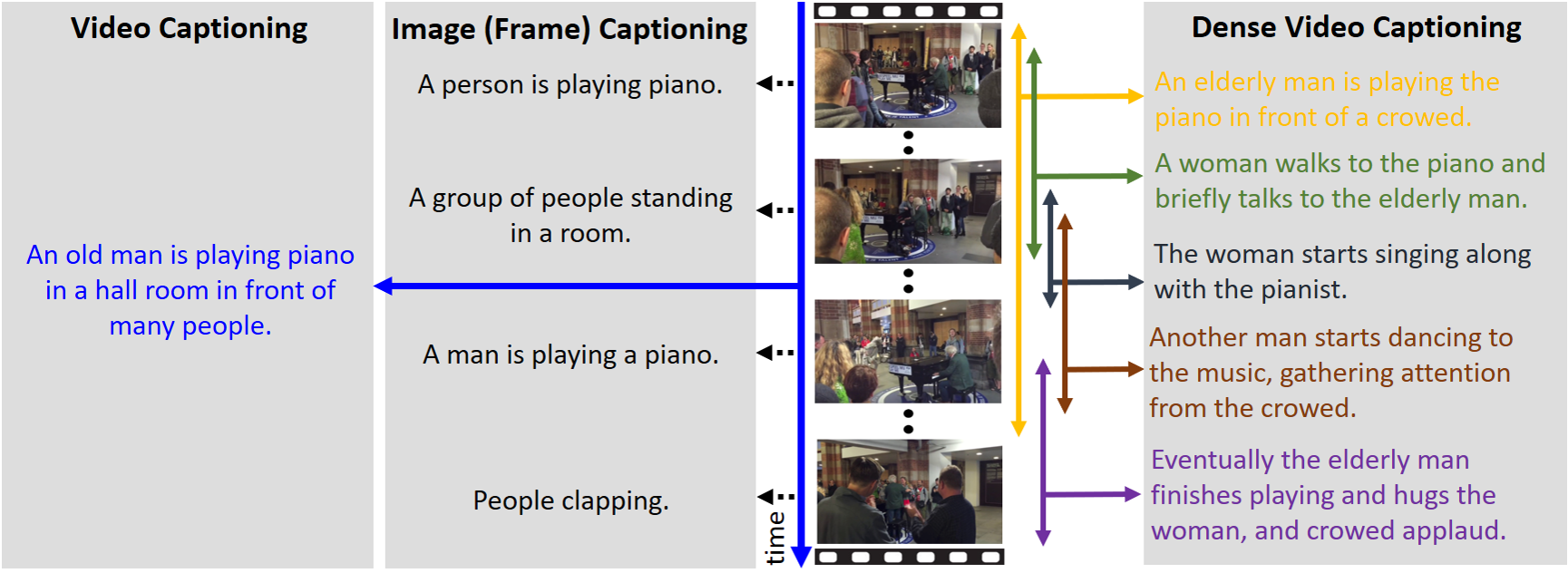} 
     \vspace{-6mm}
     \caption{Illustration of differences between image captioning, video captioning and dense video captioning. Image (video frame) captioning describes each frame with a single sentence. Video captioning describes the complete video with one sentence. In dense video captioning, each event in video is temporally detected and described by a single sentence eventually resulting in multiple sentences localized in time but not necessarily coherent.}
\label{fig:piano}
\vspace{-5mm}
\end{figure*}

Automatically generating natural language sentences describing the video content has two components; \emph{understanding} the visual content and \emph{describing} it in grammatically correct natural language sentences. Figure \ref{fig:basicblockdiagram} shows a simple deep learning based video captioning framework. The task of video description is relatively more challenging, compared to image captioning, because not all objects in the video are relevant to the description such as the detected objects that do not play any role in the observed activity~\cite{barbu2012video}. Moreover, video description methods must additionally capture the speed, direction of relevant objects as well as causality among events, actions, and objects. Finally, events in videos can be of varying lengths and may even result in a possible overlap of events~\cite{krishna2017dense}. See Figure~\ref{fig:piano} for example. The event of piano recitals is spanned over almost the entire duration of the video, however, the applause is a very short event that only takes place at the end. The example illustrates differences between three related areas of research, namely, image captioning, video captioning and dense video captioning. In this example, image captioning techniques recognize the event as mere \textit{clapping} whereas it is actually an \textit{applause} that resulted from a previous event - piano playing.

\begin{figure}[htbp] 
   \centering
   \includegraphics[width=\columnwidth]{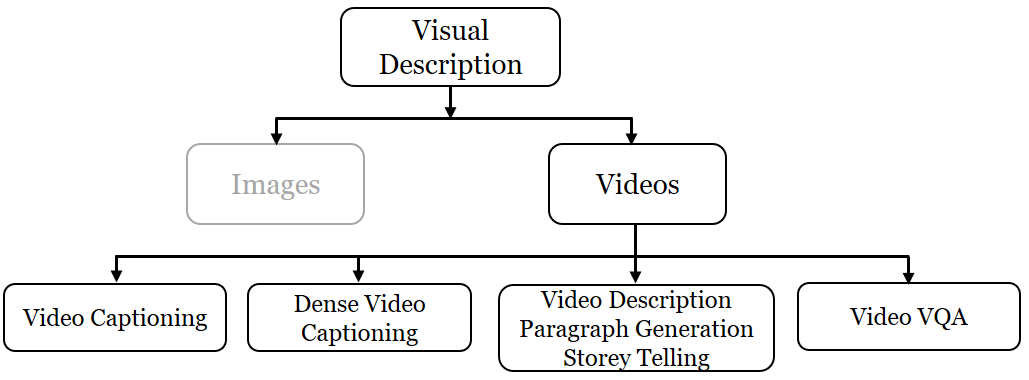} 
  \vspace{-4mm}
  \caption{Classification of visual content description. This survey focuses on video only and not images.}  
   \label{fig:captioningmodules}
\vspace{-4mm}
\end{figure}

Figure~\ref{fig:captioningmodules} summarizes related research under the umbrella of \textit{Visual Description}. The classification is based on whether the input is still images (\textit{Image Captioning}) or multi-frame short videos (\textit{Video Captioning}). Note, however, that short video captioning is very different from video auto-transcription where audio and speeches are the main focus. Video captioning concerns mainly the visual content as opposed to the audio signals. In particular, \textit{Video Description} extends video captioning with the aim to provide a more detailed account of the visual contents in the video. 

Below we define some terminologies used in this paper.

\begin{itemize}
\item {\it Visual Description}: The unifying concept encompassing (see Fig.~\ref{fig:captioningmodules}) the automatic generation of single or multiple natural language sentences that convey the information in still images or video clips.


\item {\it Video Captioning}: Conveying the information of a video clip as a whole through a single automatically generated natural language sentence based on the premise that short video clips usually contain one main event~\cite{gan2017semantic, yao2015describing, venugopalan2015sequence, ballas2015delving, Pan_2017_CVPR, donahue2015long}. 

\item {\it Video Description}: Automatically generating multiple natural language sentences that provide a narrative of a relatively longer video clip. The descriptions are more detailed and may be in the form of paragraphs. Video description is sometimes also referred to as \textit{story telling} or \textit{paragraph generation}~\cite{yu2016video, rohrbach2014coherent}.


\item {\it Dense Video Captioning}: Detection and conveying information of all, possibly overlapping, events of different lengths in a video using a natural language sentence per event. As illustrated in Fig.~\ref{fig:piano}, dense video captioning localizes events in time~\cite{krishna2017dense, yaomsr, ruccmu, xu2018joint} and generates sentences that are not necessarily coherent. On the other hand video description gives a more detailed account of one or more events in a video clip using multiple coherent sentences without having to localize individual events.
\end{itemize}

Video captioning research started with the classical template based approaches in which Subject (S), Verb (V), and Object (O) are detected separately and then joined using a sentence template. These approaches are referred to as \textit{SVO-Triplets} \cite{kojima2002natural,barbu2012video}. However, the advent of deep learning and the tremendous advancements in CV and NLP have equally affected the area of video captioning.
Hence, latest approaches follow deep learning based architectures \cite{venugopalan2015sequence,rohrbach2017movie} that encode the visual features with 2D/3D-CNN and use LSTM/GRU to learn the sequence.
The output of both approaches is either a single sentence \cite{xu2016msr,pan2016jointly}, or multiple sentences \cite{rohrbach2014coherent,barbu2012video, shin2016beyond, yu2016video, das2013thousand, khan2011human} per video clip. 
Early research on video description mostly focused on domain specific short video clips with limited vocabularies of objects and activities \cite{barbu2012video,das2013thousand,kojima2002natural,khan2012describing,rohrbach2013translating,yu2013grounded}. Description of open domain and relatively longer videos remains a challenge, as it needs large vocabularies and training data. 
  Methods that follow CNN-LSTM/GRU framework mainly differ from each other in the different types of CNNs and language models (vanilla RNN, LSTM, and GRUs) they employ and as well as how they pass the extracted visual features to the language model (at the first time step only or all time steps). Later methods progressed by introducing additional transformations on top of the standard encoder-decoder framework. These transformations include attention mechanism~\cite{yao2015describing} where the model learns which part of the video to focus on, sequence learning~\cite{venugopalan2015sequence} that models a sequence of video frames with the sequence of words in the corresponding sentence, semantic attributes~\cite{gan2017semantic, Pan_2017_CVPR} that exploits the visual semantics in addition to CNN features, and joint modeling of visual content with compositional text~\cite{pan2016jointly}. More recently, video based visual description problem has evolved towards dense video captioning and video story telling. New datasets have also been introduced to progress along these lines. 
  
When it comes to performance comparison, quantitative evaluation of video description systems is not straightforward. Currently, automatic evaluations are typically performed using machine translation and image captioning metrics, including Bilingual Evaluation Understudy (BLEU)~\cite{papineni2002bleu}, Recall Oriented Understudy for Gisting Evaluation (ROUGE)~\cite{lin2004rouge}, Metric for Evaluation of Translation with Explicit Ordering (METEOR)~\cite{lavie2005meteor}, Consensus based Image Description Evaluation (CIDEr)~\cite{vedantam2015cider}, and the recently proposed Semantic Propositional Image Captioning Evaluation (SPICE)~\cite{anderson2016spice} and Word Mover's Distance (WMD)~\cite{kusner2015word} metrics. Section \ref{sec:evalmetricsintro} presents these measures. Here, we give a brief overview to establish motivation for our survey. BLEU is a precision-based metric, which accounts for precise matching of \textit{n-grams} in the generated and ground truth references. METEOR, on the other hand, first creates an alignment between the two sentences by comparing exact tokens, stemmed tokens and paraphrases. It also takes into consideration the semantically similar matches using WordNet synonyms. ROUGE, similar to BLEU, has different \textit{n-grams} based versions and computes recall for the generated sentences and the reference sentences. CIDEr is a human-consensus-based evaluation metric, which was developed specifically for evaluating image captioning methods but has also been used in video description tasks. WMD makes use of word embeddings (semantically meaningful vector representations of words) and compares two texts using the Earth Mover's Distance (EMD). This metric is relatively less sensitive to word order and synonym changes in a sentence and, like CIDEr and METEOR, it provides high correlation with human judgments. Lastly, SPICE is a more recent metric that correlates more with human judgment of semantic quality as compared to previously reported metrics. It compares the semantic information of two sentences by matching their content in dependency parse trees. These metrics capture very different performance measures for the same method and are not perfectly aligned with human judgments. 
Also, due to the hand engineered nature of these metrics, their scores are unstable when the candidate sentence is perturbed with synonyms, word order, length and redundancy. Hence, there is a need for an evaluation metric that is {\em learned} from training data to score in harmony with human judgments in describing videos with diverse content.

The current literature lacks a comprehensive and systematic survey that covers different aspects of video description research including methods, dataset characteristics, evaluation measures, benchmark results and related competitions and video Q\&A challenges. We fill this gap and present a comprehensive survey of the literature. We first highlight the important applications and major trends of video description in Section \ref{sec:introduction} and then classify automatic video description methods into three groups, giving an overview of the models from each group in Section~\ref{sec:Methods}. In Section \ref{ch:datasetssection}, we elaborate on the available video description datasets used for benchmarking. In Section \ref{videocompetitions}, we present the details of video competitions and challenges. Furthermore, we review the evaluation metrics that are used for quantitative analysis of the generated descriptions in Section \ref{sec:EvalMetPreamble}. In Section \ref{ch:results}, benchmark results achieved through the aforementioned methods are compared and discussed. In Section~\ref{sec:futuredirections}, we discuss the possible future directions and finally Section \ref{sec:conc} concludes our survey and discusses some insights into the findings.
\vspace{-3mm}
\section{Video Description Methods}
\label{sec:Methods}
Video description literature can be divided into three main phases. The classical methods phase, where pioneering visual description research employed classical CV and NLP methods to first detect entities (objects, actions, scenes) in videos and then fit them to standard sentence templates. The statistical methods phase, which employed statistical methods to deal with relatively larger datasets. This phase lasted for a relatively short time.  Finally, the deep learning phase, which is the current state of the art and is believed to have the potential to solve the open domain automatic video description problem. Below, we give a detailed survey of the methods in each category.

\subsection{Classical Methods} 
\label{sec:svo}
The SVO (Subject, Object, Verb) tuples based methods are among the first successful methods used specifically for video description. 
However, research efforts were made long before to describe visual content into natural language, albeit not explicitly for captioning or description. The first ever attempt goes back to Koller et al.~\cite{koller1991algorithmic} in 1991, who developed a system that was able to characterize motion of vehicles in real traffic scenes using natural language verbs. Later in 1997, Brand et al.~\cite{brand1997inverse} dubbed this as "Inverse Hollywood Problem" (since in Hollywood script (description) is converted into video, here the problem is opposite), and described a series of actions into semantic tag summaries in order to develop a storyboard from instructional videos. They also developed a system, ``video gister'', that was able to heuristically parse the videos into a series of key actions and generate a script that describes actions detected in the video. They also generated key frames depicting the detected causal events and defined the series of events into semantics representation e.g. {\tt Add by enter, motion, detach and remove by attach, move, leave.} Video gister was limited to only one human arm (actor) interacting with non liquid objects and was able to understand only five actions (touch, put, get, add, remove).


Getting back to SVO tuple based methods, which tackle the video description generation task in two stages. The first stage known as \textit{content identification} focuses on visual recognition and classification of the main objects in the video clip. These typically include the performer or \textit{actor}, the \textit{action} and the \textit{object} of that action. The second stage involves \textit{sentence generation} which maps the objects identified in the first stage to Subject, Verb and Object (and hence the name SVO), and filling in handcrafted templates for grammatically sound sentences. These templates are created using grammar or rule-based systems, which are only effective in very constrained environments, i.e. short clips or videos with limited number of objects and actions. 

Numerous method have been proposed for detecting objects, humans, actions, and events in videos. Below we summarize the recognition techniques used in the Stage I of the SVO tuples based approaches.

\begin{itemize}

\item \textit{Object Recognition:} Object recognition in SVO approaches was performed typically using conventional methods, including model-based shape matching through edge detection or color matching~\cite{kojima2002natural}, HAAR features matching~\cite{viola2001rapid}, context-based object recognition~\cite{torralba2003context}, Scale Invariant Feature Transform (SIFT)~\cite{lowe1999object}, discriminatively trained part-based models~\cite{felzenszwalb2010a} and Deformable Parts Model (DPM)~\cite{felzenszwalb2008, felzenszwalb2010b}. 

\item \textit{Human and Activity Detection: } Human detection methods employed features such as Histograms of Oriented Gradient (HOG)~\cite{dalal2005histograms} followed by SVM. For activity detection, features like Spatiotemporal Interest Points such as Histogram of Oriented Optical Flow (HOOF) \cite{hoof}, Bayesian Networks (BN)~\cite{hongeng2000bayesian}, Dynamic Bayesian Networks (DBNs)~\cite{gong2003recognition}, Hidden Markov Models (HMM)~\cite{bobick1997state}, state machines~\cite{koller1991algorithmic}, and PNF Networks~\cite{pinhanez1998human} have been used by SVO approaches. 

\item \textit{Integrated Approaches:} Instead of detecting the description-relevant entities separately, Stochastic Attribute Image Grammar (SAIG)~\cite{zhu2007stochastic} and Stochastic Context Free Grammars (SCFG)~\cite{moore2002recognizing}, allow for compositional representation of visual entities present in a video, an image or a scene based on their spatial and functional relations. Using the visual grammar, the content of an image is first extracted as a parse graph. A parsing algorithm is then used to find the best scoring entities that describe the video. In other words, not all entities present in a video are of equal relevance, which is a distinct feature of this class of methods compared to the aforementioned approaches.
\end{itemize}

For Stage II, sentence generation, a variety of methods have been proposed including HALogen representation~\cite{langkildehalogen}, Head-driven Phrase Structure Grammar (HPSG)~\cite{pollard1994head}, planner and surface realizer~\cite{reiter2000building}. The primary common task of these methods is to define templates. A template is a user-defined language structure containing placeholders. In order to function properly, a template comprises of three parts named lexicons, grammar and template rules. \textit{Lexicon} represents vocabulary that describes high level video features. \textit{Template rules} are user-defined rules guiding the selection of appropriate lexicons for sentence generation. \textit{Grammar} defines linguistic rules to describe the structure of expressions in a language, ensuring that a generated sentence is syntactically correct. Using production rules, Grammar can generate a large number of various configurations from a relatively small vocabulary.   

\begin{figure}[t] 
   \centering
   \includegraphics[width=\columnwidth]{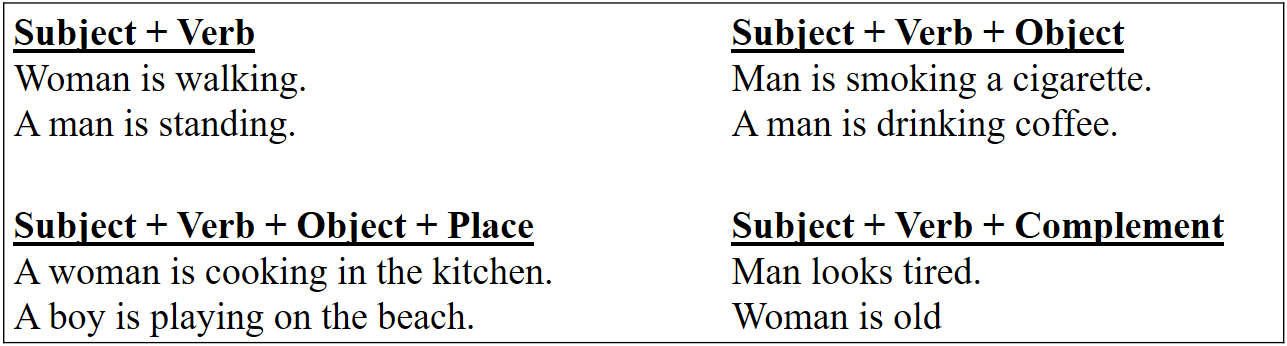} 
   \setlength{\belowcaptionskip}{-10pt}
   \caption{An example of various templates used for sentence generation from videos. Subject, verb, and object  are used to fill in these template. Verb is obtained from action/activity detection methods using spatio-temporal features whereas subject and object are obtained from object detection methods using spatial features.}
\label{fig:template}
\end{figure}

\begin{figure*}[t] 
     \centering
     \includegraphics[width=\textwidth]{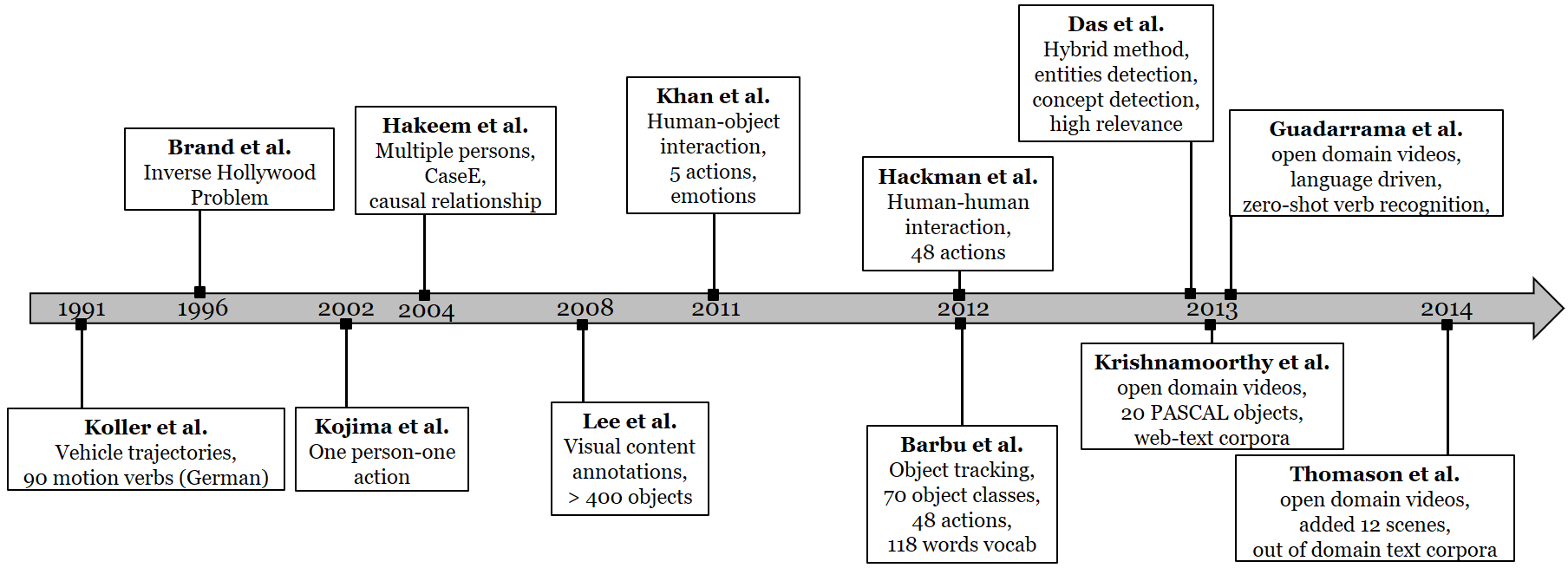} 
     \vspace{-6mm}
     \caption{Evolution of classical methods over time. In general the focus of these methods moved from subjects (humans) to actions and objects and then to open domain videos containing all three SVO categories.}
  \label{fig:classicalevolutiondia}
  \vspace{-4mm}
  \end{figure*}

In template based approaches, a sentence is generated by fitting the most important entities to each of the categories required by the template, e.g. subject, verb, object, and place. Entities and actions recognized in the content identification stage are used as lexicons. Correctness of the generated sentence is ensured by Grammar. Figure~\ref{fig:template} presents examples of some popular templates used for sentence generation in template based approaches. 
Figure \ref{fig:classicalevolutiondia} gives a timeline of how the classical methods evolved over time whereas below we provide a survey of SVO methods by grouping them into three categories namely, subject (human) focused, action and object focused and methods that use the SVO approach on open domain videos. Note that the division boundaries are frequently blurred between these categories. 

\vspace{2mm}
\noindent {\bf \underline{(1) Subject (Human) Focused:}} In 2002, Kojima et al.~\cite{kojima2002natural} proposed one of the earliest methods designed specifically for video captioning. This method focuses primarily on describing videos of one person performing one action only. To detect humans in a scene, they calculated the probability of a pixel coming from the background or the skin region using the values and distributions of pixel chromaticity. Once a human's head and hands are detected, the human posture is estimated by considering three kinds of geometric information i.e. position of the head and hands and direction of the head. For example, to obtain the head direction, the detected head image is compared against a list of pre-collected head models and a threshold is used to decide on the matching head direction. For object detection, they applied two-way matching, i.e. shape-based matching and pixel based color matching to a list of predefined known objects. Actions detected are all related to object handling and the difference image is used to detect actions such as putting an object down or lifting an object up. To generate the description in sentences, pre-defined {\it case frames} and verb patterns as proposed by Nishida et al.~\cite{nishida1988feedback, nishida1982japanese} are used. Case frame is a type of frame expression used for representing the relationship between cases, which are classified into 8 categories. The frequently used ones are \textit{agent}, \textit{object}, and \textit{locus}. For example, ``a person walks from the table to the door'', is represented as:

{\tt[PRED:walk, AG:person, GO-LOC:by(door), SO-LOC:front(table)]},

\noindent
where {\tt PRED} is the predicate for action, {\tt AG} is the agent or actor, {\tt GO-LOC} is the goal location and {\tt SO-LOC} is the source location. A list of semantic primitives are defined about movements, which are organized using body action state transitions. For example, if {\tt moving} is detected and the speed is {\tt fast}, then the activity state is transitioned from {\tt moving} to {\tt running}. They also distinguish durative actions (e.g. walk) from instantaneous actions (e.g. stand up). The major drawback of their approach is that it cannot be easily extended to more complex scenarios such as multiple actors, incorporating temporal information, and capturing causal relationship between events. The heavy reliance on the correctness of manually created activity concept hierarchy and state transition model also prevents it from being used in practical situations. 

Hakeem et. al.~\cite{hakeem2004case} addressed the shortcomings of Kojima et. al's~\cite{kojima2002natural} work and proposed an extended case framework ({\tt CASE$^E$}) using hierarchical {\tt CASE} representations. They incorporated multiple agent events, temporal information, and causal relationship between the events to describe the events in natural language. They introduced case-list to incorporate multiple agents in {\tt AG}, {\tt[PRED:move, AG:\{person1, person2\},...]}. Moreover, they incorporated temporal information into CASE using temporal logic to encode the relationship between sub-events. As some events are conditional on other events, they also captured causal relationship between events. For example, in the sentence "a man played piano and the crowd applauded", the applaud occurred because the piano was played. 
{\tt[CAUSE: [PRED:play, D:crowed, FAC:applaud]]}. 

Khan et al.~\cite{khan2011human} introduced a framework to describe human related contents such as actions (limited to five only) and emotions in videos using natural language sentences. They implemented a suite of conventional image processing techniques, including face detection~\cite{kuchi2002human}, emotion detection~\cite{maglogiannis2009face}, action detection~\cite{bobick1997state}, non-human object detection~\cite{viola2001rapid} and scene classification~\cite{kim2010novel}, to extract the high level entities of interest from video frames. These include humans, objects, actions, gender, position and emotion. Since their approach encapsulates human related actions, human is rendered as \textit{Subject} and the objects upon which action is performed are rendered as \textit{Object}. A template based approach is adopted to generate natural language sentences based on the detected entities. They evaluated the method on a dataset of 50 snippets, each spanning 5 to 20 seconds duration. Out of 50, 20 snippets were human close-ups and 30 showed human activities such as stand, walk, sit, run and wave. The primary focus of their research was on activities involving a human interacting with some objects. Hence, their method does not generate any  description until a human is detected in the video. The method cannot identify actions with subtle movements (such as smoking and drinking) and interactions among humans.

\vspace{2mm}
\noindent {\bf \underline{(2) Action and Object Focused:}} Lee et al.~\cite{lee2008save} proposed a method for semantically annotating visual content in three sequential stages namely, image parsing, event inference and language generation. An ``image parsing engine'' using  stochastic attribute image grammar (SAIG)~\cite{zhu2007stochastic} is employed to produce a visual vocabulary i.e. a list of visual entities present in the frame along with their relationships. This output is then fed into an ``event inference engine'', which extracts semantic and contextual information of visual events, along with their relationships. Video Event Markup Language (VEML)~\cite{nevatia2004ontology} is used to represent semantic information. In the final stage, head-driven phrase structure grammar (HPSG)~\cite{pollard1994head} is used to generate text description from the semantic representation. Compared to Kojima et al.~\cite{kojima2002natural}, grammar-based methods can infer and annotate a wider range of scenes and events. Ten streams of urban traffic and maritime scenes over a period of 120 minutes, containing more than 400 moving objects are used for evaluation. Some detected events include ``entering  the scene, moving, stopping, turning, approaching traffic intersection, watercraft approaching maritime markers and land areas and scenarios where one object follows the other''~\cite{lee2008save}. Recall and Precision rates are employed to evaluate the accuracy of the events that are detected with respect to manually labeled ground truth. Due to poor estimation of the motion direction from low number of perspective views, their method does not perform well on ``turning'' events.

Hanckmann et al.~\cite{hanckmann2012automated} proposed a method to automatically describe events involving multiple actions (7 on average), performed by one or more individuals. Unlike Khan et al.~\cite{khan2011human}, human-human interactions are taken into account in addition to human-object interactions. Bag-of-features (48 in total) are collected as action detectors~\cite{burghouts2012recognition} for detecting and classifying actions in a video. The description generator subsequently describes the verbs relating the actions to the scene entities. It finds the appropriate actors among objects or persons and connects them to the appropriate verbs. In contrast to Khan et al.~\cite{khan2011human} who assume that the subject is always a person, Hanckmann et al.~\cite{hanckmann2012automated} generalizes subjects to include vehicles as well. Furthermore, the number of human actions is much richer. Compared to the five verbs in Khan et al.~\cite{khan2011human}), they have 48 verbs capturing  a diverse range of actions such as \texttt{approach}, \texttt{arrive}, \texttt{bounce}, \texttt{carry}, \texttt{catch} and etc.

Barbu et al. \cite{barbu2012video} generated sentence descriptions for short videos of highly constrained domains consisting of 70 object classes, 48 action classes and a vocabulary of 118 words. They rendered a detected object and action as noun and verb respectively. Adjectives are used for the object properties and prepositions are used for their spatial relationships. Their approach comprises of three steps. In the first step, object detection~\cite{felzenszwalb2010b} is carried out on each frame by limiting 12 detections per frame to avoid over detections. Second, object tracking~\cite{tomasi1991detection, shi1994good} is performed to increase the precision. Third, using dynamic programming the optimal set of detections is chosen. Verb labels corresponding to actions in the videos are then produced using Hidden Markov Models (HMMs). After getting the verb, all tracks are merged to generate template based sentences that comply to grammar rules. 

\begin{figure*}[htbp]
   \centering
   \includegraphics[width=0.9\textwidth]{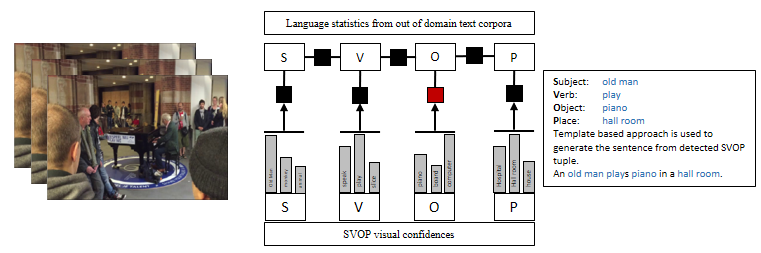} 
   \vspace{-3mm}
   \caption{Example of the Subject-Verb-Object-Place (SVOP) \cite{thomason2014integrating} approach where confidences are obtained by integrating probabilities from visual recognition system, with statistics from out of domain English text corpora to determine the most likely SVOP tuple. The red block shows low probability given to a correct object by the visual system that is rectified by the high probability from the linguistic model.}
\label{fig:svoexample}
    \vspace{-5mm}
\end{figure*}

Despite the reasonably accurate lingual descriptions generated for videos in constrained environments, the aforementioned methods have trouble scaling to accommodate increased number of objects and actions in open domain and large video corpora. To incorporate all the relevant concepts, these methods require customized detectors for each entity. Furthermore, the texts generated by existing methods of the time have mostly been in the form of putting together lists of keywords using grammars and templates without any semantic verification. To address the issue of lacking semantic verification, Das et. al~\cite{das2013thousand} proposed a hybrid method that produces content of high relevance compared to simple keyword annotation methods. They borrowed ideas from image captioning techniques. 
This hybrid model comprises of three steps in a hierarchical manner. 
First, in a bottom up approach, keywords are predicted using low level video features. In this approach they first find a proposal distribution over the training set of vocabulary using \textit{multimodal latent topic models}. Then by using grammar rules and parts of speech (POS) tagging, most probable subjects, objects and verbs are selected. Second, in a top down approach, a set of concepts is detected and stitched together. A tripartite graph template is then used for converting the stitched concepts to a natural language description. Finally, for semantic verification, they produced a ranked set of natural language sentences by comparing the predicted keywords with the detected concepts. Quantitative evaluation of this hybrid method shows that it was able to generate more relevant content compared to its predecessors~\cite{barbu2012video,khan2012describing}.

\vspace{2mm}
\noindent {\bf \underline{(3) SVO Methods for Open Domain Videos:}} While most of the prior mentioned works are restricted to constrained domains, Krishnamoorthy et al.~\cite{krishnamoorthy2013generating} lead the  early works of describing open domain videos. They used selected open domain YouTube videos, however, the subjects and objects were limited to the 20 entities that were available in the classifier training set. Their main contribution is the introduction of text-mining using web-scale text corpora to aid the selection of the best SVO tuple to improve sentence coherence. 

In addition to focusing on open domain videos and utilizing web scaled text corpora, Guadarrama et al.~\cite{guadarrama2013youtube2text} and Thomason et al.~\cite{thomason2014integrating} started dealing with relatively larger vocabularies. Compared to Krishnamoorthy et al.~\cite{krishnamoorthy2013generating}, instead of using only 20 objects in the PASCAL dataset~\cite{everingham2010pascal}, all videos of the YouTube corpora are used for the detection of 241 objects, 45 subjects, and 218 verbs. To describe short YouTube videos, Guadarrama et al.~\cite{guadarrama2013youtube2text} proposed a novel language driven approach. They introduced ``zero-shot" verb recognition for selecting unseen verbs in the training set. For example, if subject is ``person", object refers to ``car" and the model-predicted verb is ``move", then the most suitable verb would be ``drive". Thomason et al. \cite{thomason2014integrating} used visual recognition techniques on YouTube videos for probabilistic estimations of subjects, verbs, and objects. Their approach is illustrated in Figure~\ref{fig:svoexample}. The object and action classifiers were trained on ImageNet~\cite{russakovsky2015imagenet}. In addition to detecting subjects, verbs and objects, places (12 scenes) where actions are performed, e.g. kitchen or play ground are also identified. To further improve the accuracy of assigning visually detected entities to the right category, probabilities using language statistics obtained from four ``out of domain" English text corpora: English Gigaword, British National Corpus (BNC), ukWac and WaCkypedia EN are used to enhance the confidence of word-category alignment for sentence generation. A small ``in domain" corpus comprising human-annotated sentences for the video description dataset is also constructed and incorporated into the sentence generation stage. Co-occurring bi-gram (SV, VO, and OP) statistics from the candidate SVOP tuples are calculated using both the ``out of domain" and the ``in domain" corpus, which are used in a Factor Graph Model (FGM) to predict the most probable SVO and place combination. Finally, the detected SVOP tuple is used to generate an English sentence through a template based approach. 

Classical methods focused mainly on the detection of pre-defined entities and events separately. These methods then tried to describe the detected entities and events using template based sentences. However, to describe open domain videos or those with more events and entities, classical methods must employ object and action detection techniques for each entity which is unrealistic due to the computational complexity. Moreover, template based descriptions are insufficient to describe all possible events in videos given the linguistic complexity and diversity. Consequently, these methods failed to describe semantically rich videos.

\begin{figure*}[htbp] 
   \centering
   \includegraphics[width=0.7\textwidth]{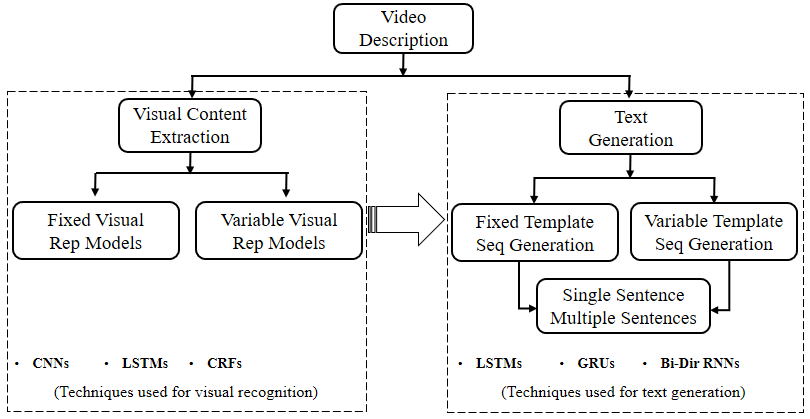} 
   \vspace{-2mm}
   \caption{Deep learning based video description techniques in the literature comprise two main stages. The first stage involves visual content extraction and is represented either by a fixed length vector or by dynamic vectors. The second stage takes input of visual representation vectors from the first stage for text generation and generates single/multiple sentence(s).}
   \label{fig:blockdia}
   \vspace{-3mm}
\end{figure*} 

\subsection{Statistical Methods}
\label{sec:statisticalmethods}
Na\"ive SVO tuple rule-based engineering approaches are indeed inadequate to describe open domain videos and large datasets, such as YouTubeClips~\cite{chen2010collecting}, TACoS-MultiLevel \cite{rohrbach2014coherent}, MPII-MD~\cite{rohrbach2015dataset}, and M-VAD~\cite{torabi2015using}.
These datasets contain very large vocabularies as well as tens of hours of videos. There are three important differences between 
these open domain and previous datasets. Firstly, open domain videos contain unforeseeable diverse set of subjects, objects, activities and places. Secondly, due to the sophisticated nature of human languages, such datasets are often annotated with multiple viable meaningful descriptions. Thirdly, the videos to be described are often long, potentially stretching through many hours. Descriptions of such videos with multiple sentences or even paragraphs become more desirable.
 
To avoid the tedious efforts required in rule-based engineering methods, Rohrbach et. al.~\cite{rohrbach2013translating} proposed a machine learning method to convert visual content into natural language. They used parallel corpora of videos and associated annotations. Their method follows a two step approach. First, it learns to represent the video as intermediate semantic labels using maximum posterior estimate (MAP). Then, it translates the semantic labels into natural language sentences by using techniques borrowed from Statistical Machine Translation (SMT)~\cite{koehn2007moses}.
In this machine translation approach, the intermediate semantic label representation is the source while the expected annotations are regarded as the target language.

For the object and activity recognition stages, the research moved from earlier threshold-based detection~\cite{kojima2002natural} to manual feature engineering and traditional classifiers~\cite{krishnamoorthy2013generating,das2013thousand,guadarrama2013youtube2text, thomason2014integrating}. For the sentence generation stage, an uptake of machine learning methods can be observed in recent years to address the issue of large vocabulary. This is also evidenced by the trend in recent methods that use models for lexical entries that are learned 
in a weakly supervised \cite{rohrbach2014coherent,rohrbach2013translating,xu2015jointly,yu2015learning} or fully supervised \cite{corso2015gbs,guadarrama2013youtube2text,krishnamoorthy2013generating,sun2014semantic} fashion. However, the separation of the two stages makes this camp of methods incapable of capturing the interplay of visual features and linguistic patterns, let alone learning a transferable state space between visual artifacts and linguistic representations. In the next section, we review the deep learning methods and discuss how they address the scalability, language complexity and domain transferability issues faced by open domain video description.

\begin{figure*}[htbp] 
   \centering
   \includegraphics[width=\textwidth]{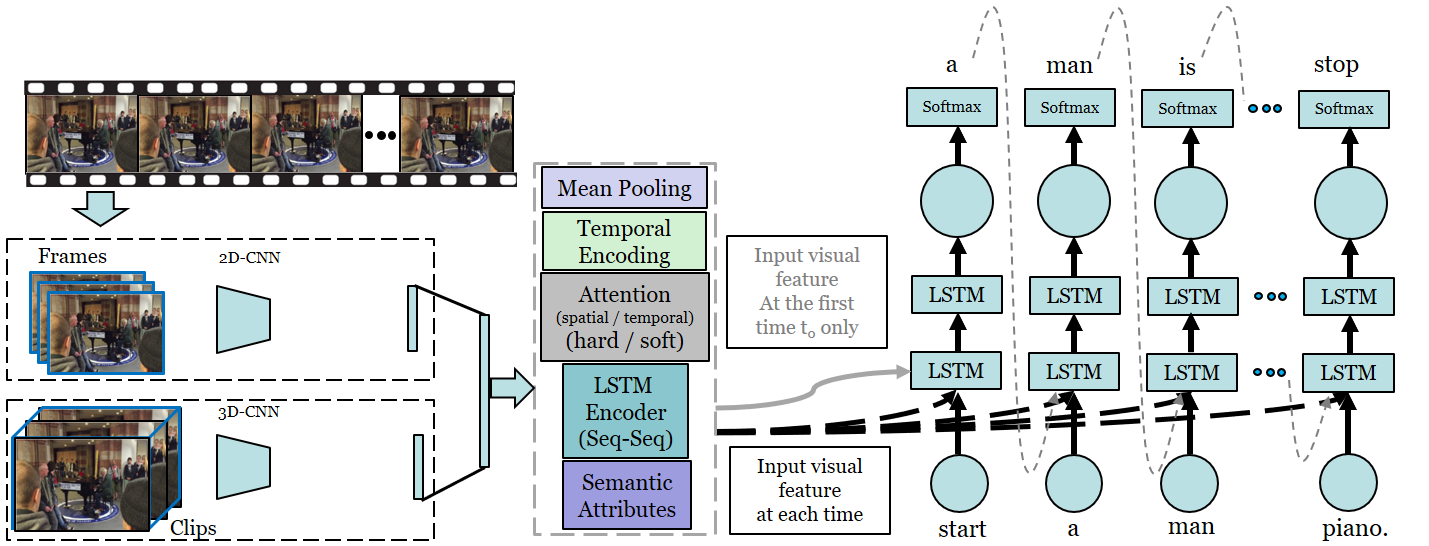} 
   \caption{Summary of deep learning based video description methods. Most methods employ mean pooling of frame representations to represent a video. More advanced methods use attention mechanisms, semantic attribute learning, and/or employ a sequence-to-sequence approach. These methods differ in whether the visual features are fed only at first time step or all time steps of the language model.}
\label{fig:methodssummary}
\vspace{-3mm}
\end{figure*}

\subsection{Deep Learning Models}       

The whirlwind success of deep learning in almost all sub-fields of computer vision, has also revolutionized video description approaches. In particular, Convolutional Neural Networks (CNNs)~\cite{krizhevsky2012imagenet} are the state of the art for modeling visual data and excel at tasks such as object recognition~\cite{krizhevsky2012imagenet, simonyan2014very, szegedy2015going}. Long Short-Term Memory (LSTMs)~\cite{hochreiter1997long} and the more general deep Recurrent Neural Networks (RNNs), on the other hand, are now dominating the area of sequence modeling, setting new benchmarks in machine translation \cite{sutskever2014sequence, cho2014properties}, speech recognition \cite{graves2014towards} and the closely related task of image captioning \cite{donahue2015long, vinyals2015show}. While conventional methods struggle to cope with large-scale, more complex and diverse datasets for video description, researchers have combined these deep nets in various configurations with promising performances.

As shown in Figure~\ref{fig:blockdia}, the deep learning approaches to video description can also be divided into two sequential stages, namely, visual content extraction and text generation. However, in contrast to the SVO Tuple Methods in Section~\ref{sec:svo}, where lexical word tokens are generated as a result of the first stage through visual content extraction, visual features represented by fixed or dynamic real-valued vectors are produced instead. This is often referred to as the {\it video encoding stage}.  CNN, RNN or Long Short-Term Memory (LSTM) are used in this encoding stage to learn these visual features, that are then used in the second stage for text generation, also known as the {\it decoding stage}. For decoding, different flavours of RNNs are used, such as deep RNN, Bi-directional RNN, LSTM or Gated Recurrent Units (GRU). The resulting description can be a single sentence or multiple sentences. Figure~\ref{fig:methodssummary} illustrates a typical end-to-end video description system with encoder-decoder stages. The encoding part is followed by transformations such as mean pooling, temporal encoding or attention mechanisms to represent the visual content. Some methods apply sequence-to-sequence learning and/or semantic attributes learning in their frameworks. The aforementioned mechanisms have been used in different combinations by contemporary methods. We group the literature based on the different combinations of deep learning architectures for encoding and decoding stages, namely:

\begin{itemize}
\item CNN - RNN Video Description, where convolution architectures are used for visual encoding and recurrent structures are used for decoding. This is the most common architecture employed in deep learning based video description methods; 
\item RNN - RNN Video Description, where recurrent networks are used for both stages; and  
\item Deep reinforcement networks, the relatively new research area for video description.
\end{itemize}

\subsubsection{CNN-RNN Video Description}
Given its success in computer vision and simplicity, CNN is still by far the most popular network structure used for visual encoding. The encoding process can be broadly categorized into fixed-size and variable-size video encoding.

Donahue et al.~\cite{donahue2015long} were the first to use a deep neural networks to solve the video captioning problem. They proposed three architectures for video description. Their model is based on the assumption to have CRF based predictions of subjects, objects, and verbs after full pass of complete video. This allows the architecture to observe the complete video at each time step. The first architecture, LSTM encoder-decoder with CRF max, is motivated by the statistical machine translation (SMT) based video description approach by Rohrbach et al.~\cite{rohrbach2013translating} mentioned earlier in Section~\ref{sec:statisticalmethods}. Recognizing the state of the art machine translation performance of LSTMs, the SMT module in ~\cite{rohrbach2013translating} is replaced with a stacked LSTM comprising two layers for encoding and decoding. Similar to~\cite{sutskever2014sequence}, the first LSTM layer encodes the one-hot vector of the input sentence allowing for variable-length inputs. The final hidden representation from the first encoder stage is then fed into the decoder stage to generate a sentence by producing one word per time step. Another variant of the architecture, LSTM decoder with CRF max, incorporates max predictions. This architecture encodes the semantic representation into a fixed length vector. Similar to image description, LSTM is able to see the whole visual content at every time step. An advantage of LSTM is that it is able to incorporate probability vectors during training as well as testing. This virtue of LSTM is exploited in the third variant of the architecture, LSTM decoder with CRF probabilities. Instead of using max predication like in second variant (LSTM decoder with CRF max), this architecture incorporates probability distributions. Although the LSTM outperformed the SMT based approach of~\cite{rohrbach2013translating}, it was still not trainable in an end-to-end fashion.

In contrast to the work by Donahue et al.~\cite{donahue2015long}, where an intermediate role representation was adopted, Venugopalan et al.~\cite{venugopalan2014translating} presented the first end-to-end trainable network architecture for generating natural language description of videos. Their model is able to simultaneously learn the semantic as well as grammatical structure of the associated language. Moreover, Donahue et al.~\cite{donahue2015long} presented results on {\it domain specific} cooking videos comprising pre-defined objects and actors.  On the other hand, Venugopalan et al.~\cite{venugopalan2014translating} reported results on {\tt open domain} YouTube Clips~\cite{chen2011collecting}. To avoid supervised intermediate representations, they connected an LSTM directly to the output of the CNN. The CNN extracts visual features whereas the LSTM models the sequence dynamics. They transformed a short video into a fixed length visual input using a CNN model~\cite{ jia2014caffe} that is slightly different from AlexNet~\cite{krizhevsky2012imagenet}. The CNN model~\cite{ jia2014caffe} was learned using the ILSVRC-2012 object classification dataset (comprising 1.2M images), which is a subset of ImageNet~\cite{russakovsky2015imagenet}. It provides a robust and efficient way without manual feature selection for initialization object recognition in the videos. They sampled every tenth frame in the video and extracted features for all sample frames from the \textit{fc7} layer of the CNN. Furthermore, they represented a complete video by averaging all the extracted frame-wise feature vectors into a single vector. These feature vectors are then fed into a two-layered LSTM~\cite{graves2013speech}. The feature vectors from CNN form the input to the first layer of the LSTM. A second LSTM layer is stacked on top of first LSTM layer, where the hidden state of the first LSTM layer becomes the input to the second LSTM unit for caption generation. In essence, the transforming of multiple frame-based feature vectors into a single aggregated video-based vector, reduces the video description problem into an image captioning one. This end-to-end model performed better than the previous video description systems at the time and was able to effectively generate the sequence without any templates. However, as a result of simple averaging, valuable temporal information of the video, such as the order of appearances of any two objects, are lost. Therefore, this approach is only suitable of generating captions for short clips with a single major action in the clip.


Open domain videos are rich in complex interactions among actors and objects. Representation of such videos using a temporally averaged single feature vector is, therefore, prone to produce clutter. Consequently, the descriptions produced are bound to be inadequate because valuable temporal ordering information of events are not captured in the representation. With the success of C3D~\cite{c3dgenericfeatures} in capturing spatio-temporal action dynamics in videos, Li et al.~\cite{yao2015describing} proposed a novel 3D-CNN to model the spatio-temporal information in videos. Their 3D-CNN is based on GoogLeNet~\cite{szegedy2015going} and  pre-trained on an activity recognition dataset. It captures local fine motion information between consecutive frames. This local motion information is then subsequently summarized and preserved through higher-level representations by modeling a video as a 3D spatio-temporal cuboid. It is further represented by concatenation of HoG, HoF, MbH~\cite{dalal2006human, wang2009evaluation}. These transformations not only help capture local motion features but also reduce the computation of the subsequent 3D CNN. For global temporal structure, a temporal attention mechanism is proposed and adapted from soft attention~\cite{bahdanau2014neural}. Using 3D CNN and attention mechanisms in RNN, they were able to improve results. Recently, GRU-EVE~\cite{gruevehftsem} was proposed as an effective and computationally efficient technique for video captioning. GRU-EVE uses a standard GRU for language modeling but with Enriched Visual Encoding  as follows. It applies the Short Fourier Transform on 2D/3D-CNN features in a hierarchical manner to encapsulate the spatio-temporal video dynamics. The visual features are further enriched with high level semantics of the detected objects and actions in the video. Interestingly, the enriched features obtained by applying Short Fourier Transform on 2D-CNN features alone~\cite{gruevehftsem}, outperform C3D~\cite{c3dgenericfeatures} features.

Unlike the \textit{fixed video representation models} discussed above, \textit{variable visual representation models} are able to directly map input videos comprising different number of frames to variable length words or sentences (outputs), and are successful in modeling various complex temporal dynamics. Venugopalan et al.~\cite{venugopalan2015sequence} proposed an architecture to address the variable representation problem for both the input (video frames) and the output (sentence) stage. For that purpose they used a two-layered LSTM framework, where the sequence of video frames is input to the first layer of the LSTM. The hidden state of the first LSTM layer forms the input to the second layer of the LSTM. The output of the second LSTM layer is the associated caption. 
The LSTM parameters are shared in both stages. Although sequence-to-sequence learning had previously been used in machine translation~\cite{sutskever2014sequence}, this is the first method~\cite{venugopalan2015sequence} to use a sequence-to-sequence approach in video captioning. Later methods have adopted a similar framework, with minor variations including attention mechanisms~\cite{yao2015describing}, making a common visual-semantic-embedding~\cite{pan2016jointly} or using out of domain knowledge either with language models~\cite{venugopalan2016improving} or visual classifiers~\cite{rohrbach2015long}.

While deep learning has achieved much better results compared to previously used classifier based approaches, most methods aimed at producing one sentence from a video clip containing only one major event. In real-world applications, videos generally contain more than a single event. Description of such multi-events and semantically rich videos by only one sentence ends up to be overly simplified, and hence, uninformative. For example, instead of saying ``someone sliced the potatoes with a knife, chopped the onions into pieces and put the onions and potatoes into the pot", a single sentence generation method would probably say ``someone is cooking". Yu et al.~\cite{yu2016video} proposed a hierarchical recurrent neural network (h-RNN) that applies the attention mechanisms on both the temporal and spatial aspects. They focused on the sentence decoder and introduced a hierarchical framework that comprises of a sentence generator and on top of that a paragraph generator. First, a Gated Recurrent Unit (GRU) layer takes video features as input and generates a single short sentence. The other recurrent layer generates paragraphs using context and the sentence vectors obtained from the sentence generator. The paragraph generator thus captures the dependencies between sentences and generates a paragraph of sentences that are related.
Recently, Krishna et al. \cite{krishna2017dense} introduced the concept of dense-captioning of events in a video and employed action detection techniques to predict the temporal intervals. They proposed a model to extract multiple events with one single pass of a video, attempting to describe the detected events simultaneously. This is the first work of its kind detecting and describing multiple and overlapping events in a video. However, the model did not achieve significant improvement on the captioning benchmark.

\subsubsection{RNN - RNN Video Description}
Although not as popular as the CNN-RNN framework, another approach is to also encode the visual information using RNNs. Srivastava et al.~\cite{srivastava2015unsupervised} use one LSTM to extract features from video frames (i.e. encoding) and then pass the feature vector through another LSTM for decoding. They also introduced some variants of their models and predicted the future sequences from the previous frames. The authors adopted a machine translation model~\cite{sutskever2014sequence} for visual recognition but could not achieve significant improvement in classification accuracy.

Yu et al. \cite{yu2016video} proposed a similar approach and used two RNN structures for the video description task. Their configuration is a hierarchical decoder with multiple Gated Recurrent Units (GRU) for sentence generation. The output of this decoder is then fed to a paragraph generator which models the time dependencies between the sentences while focusing on linguistic aspects. The authors improved the state-of-the-art results for video description, however, their method is inefficient for videos involving fine-grained activities and small interactive objects. 

\subsubsection{Deep Reinforcement Learning Models}
Deep Reinforcement Learning (DRL) has out-performed humans in many real-word games. In DRL, artificial intelligent agents learn from the environment through trial and error and adjust learning policies purely from environmental rewards or punishments. DRL approaches are popularized by Google Deep Mind~\cite{mnih2015human, mnih2013playing} since 2013. Due to the absence of a straight forward cost function, learning mechanisms in this approach are considerably harder to devise as compared to traditional supervised techniques. Two distinct challenges are evident in reinforcement learning when compared with conventional supervised approaches: (1) The model does not have full access to the function being optimized. It has to query the function through interaction. (2) The interaction with the environment is state based where the present input depends on previous actions. 
The choice of reinforcement learning algorithms then depends on the scope of the problem at hand. For example, variants of Hierarchical Reinforcement Learning (HRL) framework have been applied to Atari games \cite{kulkarni2016hierarchical,vezhnevets2017feudal}. Similarly, different variants of DRL have been used to meet  the challenging requirements of image captioning \cite{ren2017deep} as well as video description \cite{pasunuru2017reinforced, wang2017video, li2018end, phan2017consensus, chen2018less}. 

Xwang et al.~\cite{wang2017video} proposed a fully-differentiable neural network architecture using reinforcement learning for video description. Their method follows a general encoder-decoder framework. The encoding stage captures the video frame features using ResNet-152~\cite{he2016deep}. The frame level features are processed through two stage encoder i.e. low level LSTM~\cite{schuster1997bidirectional} followed by a high level LSTM~\cite{hochreiter1997long}.
For decoding, they employed HRL to generate the word by word natural language descriptions. 
The HRL agent comprises of three components, a low level worker that accomplishes tasks as set by manager, a high level manager that sets goals and internal critic to ascertain whether the task has been accomplished or not and informs the manager accordingly to help manager update the goals. 
The process iterates till reaching the end of sentence token. This method is demonstrated to be capable of capturing more details of the video content thus generating more fine-grained descriptions. However, this method has shown very little improvement over existing baseline methods.

In 2018, Chen et al.~\cite{chen2018less} proposed a RL based model selecting \textit{key informative frames} to represent a complete video, in an attempt to minimize noise and unnecessary computations. Key frames are selected such that they maximize visual diversity and minimize the textual discrepancy. Hence, a compact subset of 6-8 frames on average can represented a full video. Evaluated against several popular benchmarks, it was demonstrated that video captions can be produced without performance degradation but at a significantly reduced computational cost. The method did not use motion features for encoding, a design trade-off between speed and accuracy. \\
DRL based methods are gaining popularity and have shown comparable results in video description. Due to their unconventional learning methodology, DRL methods are unlikely to suffer from paucity of labelled training data, hardware constraints and overfitting problems. Therefore, these methods are expected to flourish.

\section{Datasets}
\label{ch:datasetssection}
The availability of labeled datasets for video description have been the main driving forces behind the fast advancement of this research area. In this survey, we summarize the characteristics of these datasets and give an overview in Table~\ref{tab:datasets}. The datasets are categorized into four main classes namely \textit{Cooking}, \textit{Movies}, \textit{Videos in the Wild} and \textit{Social Media}. In most of the datasets, a single caption per video is assigned except for a few datasets which contain multiple sentences or even paragraphs per video snippet.

\begin{table*}[t]
\setlength\extrarowheight{3pt}
\small
\setlength\tabcolsep{1.5pt}
\centering
\caption{Standard datasets for benchmarking video description methods.}
\vspace{-4mm}
\begin{tabular}{|l|c|c|c|c|c|c|c|c|c|}

\noalign{\hrule height 2 pt}
    \hline
    \textbf{Dataset} & \textbf{Domain} & \textbf{\# \newline{} classes} & \textbf{\# \newline{} videos} & \textbf{avg\newline{} len} & \textbf{\# \newline{} clips} & \textbf{\# \newline{} sent} & \textbf{\# \newline{} words} & \textbf{vocab} & \textbf{len \newline{} (hrs)}\\
    \hline
    MSVD~\cite{chen2011collecting} & open  & 218 & 1970 & 10 sec & 1,970 & 70,028 & 607,339 & 13,010 & 5.3\\
    \hline
    MPII Cooking~\cite{rohrbach2012database} & cooking & 65 & 44 & 600 sec & - & 5,609 & - & - & 8.0 \\
    \hline
    YouCook~\cite{das2013thousand} & cooking  & 6 & 88 & - & Nil & 2,688 & 42,457 & 2,711 & 2.3 \\
    \hline
    TACoS~\cite{regneri2013grounding} & cooking & 26 & 127 & 360 sec & 7,206 & 18,227 & 146,771 & 28,292 & 15.9 \\
    \hline
    TACos-MLevel~\cite{rohrbach2014coherent} & cooking & 1 & 185 & 360 sec & 14,105 & 52,593 & 2,000 & - & 27.1 \\
    \hline
    MPII-MD~\cite{rohrbach2015dataset} & movie & - & 94 & 3.9 sec & 68,337 & 68,375 & 653,467 & 24,549 & 73.6 \\
    \hline
    M-VAD~\cite{torabi2015using} & movie & - & 92 & 6.2 sec & 48,986 & 55,904 & 519,933 & 17,609 & 84.6 \\
    \hline
    MSR-VTT~\cite{xu2016msr} & open & 20 & 7,180 & 20 sec & 10,000 & 200,000 & 1,856,523 & 29,316 & 41.2 \\
    \hline
    Charades \cite{sigurdsson2016hollywood} & human & 157 & 9,848 & 30 sec & - & 27,847 & - & - & 82.01 \\
    \hline
    VTW~\cite{zeng2016generation} & open & - & 18,100 & 90 sec & - & 44,613 & - & - & 213.2 \\
    \hline
    YouCook II~\cite{zhou2018towards} & cooking & 89 & 2,000 & 316 sec & 15.4k & 15.4k & - & 2,600 & 176.0 \\
    \hline
    ActyNet Cap~\cite{krishna2017dense} & open & - & 20,000 & 180 sec & - & 100,000 & 1,348,000 & - & 849.0 \\
    \hline
    ANet-Entities~\cite{zhou2018grounded} & social media & - & 14,281 & 180 sec & 52k & - & - & - & - \\
    \hline
    VideoStory~\cite{gella2018dataset} & social media & - & 20k & - & 123k & 123k & - & - & 396.0 \\
    \hline
\end{tabular}
\label{tab:datasets}
\vspace{-5mm}
\end{table*}

\subsection{Cooking} 
\subsubsection{MP-II Cooking} 
Max Plank Institute for Informatics (MP-II) Cooking dataset~\cite{rohrbach2012database} comprises 65 fine grained cooking activities, performed by 12 participants preparing 14 dishes such as \textit{fruit salad} and \textit{cake} etc. The data are recorded in the same kitchen with camera installed on the ceiling. The 65 cooking activities include ``wash hands'', ``put in bowl'', ``cut apart'', ``take out from drawer'' etc. When the person is not in the scene for 30 frames (one second) or is performing an activity that is not annotated, a ``background activity'' is generated. These fine grained activities, for example ``cut slices'', ``pour'', or ``spice'' are differentiated by movements with low inter-class and high intra-class variability. In total, the dataset comprises 44 videos (888,775 frames), with an average length per clip of approximately 600 seconds. The dataset spans a total of 8 hours play length for all videos, and 5,609 annotations.  

\subsubsection{YouCook}
The YouCook dataset \cite{das2013thousand} consists of 88 YouTube cooking videos of different people cooking various recipes. The background (kitchen/scene) is different in most of the videos. This dataset represents a more challenging visual problem than the MP-II Cooking~\cite{rohrbach2012database} dataset that is recorded with a fixed camera view point in the same kitchen and with the same background. The dataset is divided into six different cooking styles, for example \textit{grilling}, \textit{baking} etc. For machine learning, the training set contains 49 videos and the test set contains 39 videos. Frame wise annotations of objects and actions are also provided for the training videos. The object categories for the dataset include ``utensils'', ``bowls'' and ``food'' etc. Amazon Mechanical Turk (AMT) was employed for human generated multiple natural language descriptions of each video. Each AMT worker provided at least three sentences per video as a description, and on average 8 descriptions were collected per video. See Figure~\ref{fig:datasetsexample}(b) for example clips and descriptions.

\subsubsection{TACoS}
Textually Annotated Cooking Scenes (TACoS) is a subset of MP-II Composites~\cite{rohrbach2012script}. TACoS was further processed to provide coherent textual descriptions for high quality videos. Note that MP-II Composites contain more videos but less activities than the MP-II Cooking~\cite{rohrbach2012database}. It contains 212 high resolution videos with 41 cooking activities. Videos in the MP-II Composites dataset span over different lengths ranging from 1-23 minutes with an average length of 4.5 minutes. The TACoS dataset was constructed by filtering through MP-II Composites, while restricting to only those activities that involve manipulation of cooking ingredients, and have at least 4 videos for the same activity. As a result, TACoS contains 26 fine grained cooking activities in 127 videos. AMT workers were employed to align the sentences and associated videos for example: ``preparing carrots'', ``cutting a cucumber'' or ``separating eggs'' etc. For each video, 20 different textual descriptions were collected. The dataset comprises of 11,796 sentences containing 17,334 actions descriptions. A total of 146,771 words are used in the dataset. Almost 50\% of the words i.e. 75,210 describe the content for example nouns, verbs and, adjectives etc. These words includes a vocabulary size of 28,292 verb tokens. The dataset also provides the alignment of sentences describing activities by obtaining approximate time stamps where each activity starts and ends. Figure~\ref{fig:datasetsexample}(d) shows some example clips and descriptions.

\subsubsection{TACoS-MultiLevel}
\begin{figure*}[htbp] 
   \centering
   \includegraphics[width=\textwidth]{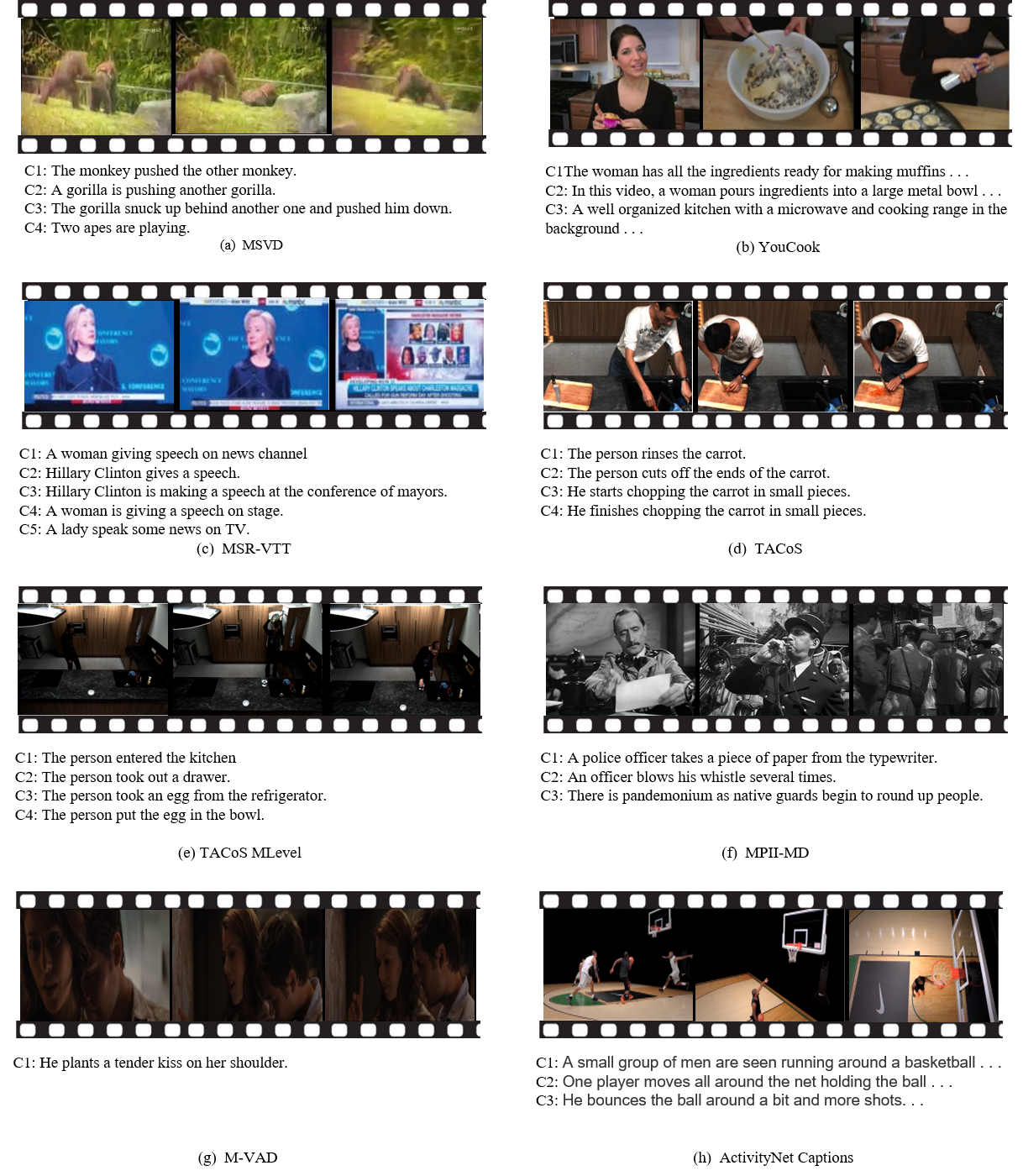} 
\vspace{-7mm}
   \caption{Example video frames (3 non-consecutive frames per clip) and captions from the various benchmark video description datasets. C1-C5 represent the associated (exemplary) captions from the dataset.}
\label{fig:datasetsexample}
\vspace{-5mm}
\end{figure*}

TACoS Multilevel \cite{rohrbach2014coherent} corpus annotations were also collected via AMT workers on the TACoS corpus \cite{regneri2013grounding}. For each video in the TACoS corpus, three levels of descriptions were collected that include: (1) detailed description of video with no more than 15 sentences per video; (2) a short description that comprises 3-5 sentences per video; and finally (3) a single sentence description of the video. Annotation of the data is provided in the form of tuples such as object, activity, tool, source and target with a person always being the subject. See Figure~\ref{fig:datasetsexample}(e) for example clips and descriptions.

\subsubsection{YouCook II}
\label{sec:youcookII}
YouCook-II Dataset~\cite{zhou2018towards} consists of 2000 videos uniformly distributed over 89 recipes. The cooking videos are sourced from YouTube and offer all challenges of open domain videos such as variations in camera position, camera motion and changing backgrounds. The complete dataset spans a total play time of 175.6 hrs and has a vocabulary of 2600 words. The videos are further divided into 3-16 segments per video  with an average of 7.7 segments per video elaborating procedural steps. Individual segment length varies from 1 to 264 seconds. All segments are temporally localized and annotated. The average length of each video is 316 seconds reaching up to a maximum of 600 seconds.  The dataset is randomly split into train, validation and test sets with the ratio of 66\%:23\%:10\% respectively.

\subsection{Movies}
\label{sec:dataset-movies}

\subsubsection{MPII-MD}
MPII-Movie Description Corpus~\cite{rohrbach2015dataset} contains transcribed audio descriptions extracted from 94 Hollywood movies. These movies are subdivided into 68,337 clips with an average length of 3.9 seconds paired with 68,375 sentences amounting to almost one sentence per clip. Every clip is paired with one sentence that is extracted from the script of the movie and the audio description data. 
The Audio Descriptions (ADs) were collected first by retrieving the audio streams from the movie using online services MakeMkV~\footnote{https://www.makemkv.com/} and Subtitle Edit~\footnote{http://www.nikse.dk/SubtitleEdit/}. These audio streams are further transcribed using crowd sourced transcription service~\cite{castingwords}. Then the transcribed texts were aligned with associated spoken sentences using their time stamps. In order to remove the misalignments of audio content with the visual content itself, each sentence was also manually aligned with the corresponding video clip. During the manual alignment process, sentences describing the content not present in the video clip were also filtered out. The audio descriptions track is an added feature in the dataset tying to describe the visual content to help visually impaired persons. The total time span of the dataset videos is almost 73.6 hours and the vocabulary size is 653,467. Example clips and descriptions are shown in Figure~\ref{fig:datasetsexample}(f).

\subsubsection{M-VAD}
Montreal Video Annotation Dataset (M-VAD)~\cite{torabi2015using} is based on the Descriptive Video Service (DVS) and contains 48,986 video clips from 92 different movies. Each clip is spanned over 6.2 seconds on average and the entire time for the complete dataset is 84.6 hours. The total number of sentences is 55,904, with few clips associated with more than one sentence. The vocabulary of the dataset spans about 17,609 words (Nouns-9,512: Verbs-2,571: Adjectives-3,560: Adverbs-857). The dataset split consists of 38,949, 4,888 and 5,149 video clips for training, validation and testing respectively. See Figure~\ref{fig:datasetsexample}(g) for example clips and descriptions.

\subsection{Social Media} 
\subsubsection{VideoStory}
\label{subsec:videostory}
VideoStory~\cite{gella2018dataset} is a multi sentence description dataset comprising 20k social media videos. This dataset is aimed to address the story narration or description generation of long videos that may not sufficiently be illustrated with single sentence. Each video is paired with at least one paragraph. The average number of temporally localized sentences per paragraph are 4.67. There are a total of 26245 paragraphs in the dataset comprising 123k sentences with an average of 13.32 words per sentence. On average, each paragraph covers 96.7\% of video content. The dataset contains about 22\% temporal overlap between co-occurring events. The dataset has training, validation and test split of 17908, 999, and 1011 videos respectively and also proposes a blind test set comprising 1039 videos. Each training video is accompanied with one paragraph, however, videos in the validation and test sets have three paragraphs each for evaluation. Annotations for the blind test are not released and are only available on server for benchmarking different methods.

\subsubsection{ActivityNet Entities}
\label{subsec:activitynetenteties}
ActivityNet Entities dataset (or ANet-Entities)~\cite{zhou2018grounded} is the first video dataset with entities grounding and annotations. This dataset is build on the training and validation splits of the ActivityNet Captions dataset~\cite{krishna2017dense}, but with different captions. In this dataset, noun phrases (NPs) of video descriptions have been grounded to bounding boxes in the video frames. The dataset comprises 14281 annotated videos, 52k video segments with at least one noun phrase annotated per segment and 158k bounding boxes with annotations. The dataset employs training set (10k) similar to ActivityNet Captions. However, validation set of ActivityNet Captions is randomly and evenly split into ANet-Entities validation (2.5k) and testing (2.5k) sets.

\subsection{Videos in the Wild}
\label{sec:datasets_general}
\subsubsection{MSVD}

Microsoft Video Description (MSVD) dataset~\cite{chen2011collecting} comprises of 1,970 YouTube clips with human annotated sentences. This dataset was also annotated by AMT workers. The audio is muted in all clips to avoid bias from lexical choices in the descriptions.  Furthermore, videos containing subtitles or overlaid text were removed during the quality control process of the dataset formulation. Finally, manual filtering was carried out over the submitted videos to ensure that each video met the prescribed criteria and was free of inappropriate and ambiguous content. The duration of each video in this dataset is typically between 10 to 25 seconds mainly showing one activity. The dataset comprises multilingual (such as Chinese, English, German etc) human generated descriptions. On average, there are 41 single sentence descriptions per clip. This dataset has been frequently used by the research community as detailed in the Results Section~\ref{ch:results}. Almost all research groups have split this dataset into training, validation and testing partitions of 1200, 100 and 670 videos respectively. Figure~\ref{fig:datasetsexample}(a) shows example clips and descriptions from MSVD dataset.

\subsubsection{MSR-VTT} 
\textit{MSR-Video to Text} (MSR-VTT)~\cite{xu2016msr} contains a wide variety of open domain videos for video captioning task. It comprises of 7180 videos subdivided into 10,000 clips. The clips are grouped into 20 different categories. An example is shown in Figure~\ref{fig:datasetsexample}(c). The dataset is divided into 6513 training, 497 validation and 2990 test videos. Each video comprises 20 reference captions annotated by AMT workers. In terms of the number of clips with multiple associated sentences, this is one of the largest video captioning datasets. In addition to video content, this dataset also contains audio information that can potentially be used for multimodal research. 

\subsubsection{Charades}
This dataset~\cite{sigurdsson2016hollywood} contains 9848 videos of daily indoor household activities. These videos are recorded by 267 AMT workers from three different continents. They were given scripts describing actions and objects and were required to follow the scripts to perform actions with the specified objects. The objects and actions used in the scripts are from a fixed vocabulary. Videos are recorded in 15 different indoor scenes and restricted to use 46 objects and 157 action classes only. The dataset comprises of 66500 annotations describing 157 actions. It also provides 41104 labels to its 46 object classes. Moreover, it contains 27847 descriptions covering all the videos. The videos in the dataset depict daily life activities with an average duration of 30 seconds. The dataset is split into 7985 and 1863 videos for training and test purposes respectively.

\subsubsection{VTW}
\textit{Video Titles in the Wild} (VTW) \cite{zeng2016generation} contains 18100 video clips with an average of 1.5 minutes duration per clip. Each clip is described with one sentence only. However, it incorporates a diverse vocabulary, where on average one word appears in not more than two sentences across the whole dataset. Besides the single sentence per video, the dataset also provides accompanying descriptions (known as augmented sentences) that describe information not present in the visual content of the clip. The dataset is proposed for video title generation as opposed to video content description but can also be used for language-level understanding tasks including video question answering. 

\subsubsection{ActivityNet Captions}
\label{subsec:activitynetcap}
ActivityNet Captions dataset~\cite{krishna2017dense} contains 100k dense natural language descriptions of about 20k videos from ActivityNet~\cite{zhu2007stochastic} that correspond to approximately 849 hours. On average, each description is composed of 13.48 words and covers about 36 seconds of video. There are multiple descriptions for every video and when combined together, these descriptions cover 94.6\% content present in the entire video. In addition, 10\% temporal overlap  makes the dataset especially interesting and challenging for studying multiple events occurring at the same time. An example of this dataset is given in Figure~\ref{fig:datasetsexample}(h).

\section{Video Description Competitions}
\label{videocompetitions}
Another major driving force of the fast-paced development in video description research comes from the many competitions and challenges organized by companies and conferences in recent years. Some of the major competitions are listed below.

\begin{table*}[htbp]
  \centering
  \small
  \caption{LSMDC Dataset Statistics.}
 \vspace{-2mm}
   \begin{tabular}{|l|c|r|r|r|r|r|}
    \hline

	\textbf{Dataset split} & \textbf{\# movies} & \textbf{\# clips} & \textbf{\# words} & \textbf{\# sent} & \textbf{avg len (sec)} & \textbf{tot len (hrs)} \\
    \hline
    LSMDC Training & 153   & 91,908 & 913,841 & 91,941 & 4.9   & 124.90 \\
    \hline
    LSMDC Validation & 12    & 6,542 & 63,789 & 6,542 & 5.2   & 9.50 \\
    \hline
    LSMDC Public Test & 17    & 10,053 & 87,147 & 10,053 & 4.2   & 11.60 \\
    \hline
    LSMDC Blind Test & 20    & 9,578 & 83,766 & 9,578 & 4.5   & 12.00 \\
    \hline
    LSMDC (Total) & 202   & 118,081 & 1,148,543 & 118,081 & 4.8   & 158.00 \\
    \hline
    \end{tabular}%
  \label{tab:LSMDC}%
\vspace{-3mm}
\end{table*}%

\subsection{LSMDC} 
The Large Scale Movie Description Challenge (LSMDC)~\cite{LSMDC2015} started in 2015 in conjunction with ICCV 2015, and as an ECCV workshop in 2016. The Challenge comprises a test set that is released publicly and a blind test set that is withheld. A server is provided to automatically evaluate~\cite{LSMDCEval} results. The challenge consists of three primary tasks i.e. {\it Movie Description}, {\it Annotation/Retrieval} and {\it Fill-in-the-Blank}. Since 2017, the {\it MovieQA} challenge has also been included in LSMDC in addition to the previous three tasks. 

The dataset for this challenge was first introduced in ICCV 2015 workshop~\cite{LSMDC2015}. The LSMDC dataset basically combines two benchmark datasets, M-VAD~\cite{torabi2015using} and MPII-MD~\cite{rohrbach2015dataset} which were initially collected independently (see Section~\ref{sec:dataset-movies}). The two datasets were merged for this Challenge, with overlaps removed to avoid repetition of the same movie in the test and training sets. Further, the manual alignments performed on MPII-MD were also removed from the validation and the test sets. The dataset was then augmented by clips only (without aligned annotations) from 20 additional movies to make up the blind test of the Challenge. These additional clips were added for evaluation only. The final LSMDC dataset has 118,081 video clips extracted from 202 unique movies. It has approximately one sentence per clip. Names of characters in the reference captions are replaced with the token word ``SOMEONE''. The dataset is further split into 91908 training clips, 6542 validations clips, 10053 public test clips and a blind (withheld) test set of 9578 clips. The average clip length is approximately 4.8 seconds. The training set captions consists of 22,829 unique words. A summary of the LSMDC dataset can be found in Table~\ref{tab:LSMDC}.

A survey of benchmark results on video description (Section-\ref{ch:results}) shows that  LSMDC has emerged as the most challenging dataset, evident by the poor performances of several models. As mentioned in the dataset section (Section~\ref{sec:dataset-movies}), natural language descriptions of movie clips are typically sourced from movie scripts and audio descriptions, so misalignments between captions and videos often occur when text refer to objects that appeared just before or after the cutting point of a clip. Misalignment is certainly a key contributing factor to the poor performances observed on this dataset. Submission protocol of the challenge is similar to the MSCOCO Image Captioning Challenge~\cite{chen2015microsoft}, and uses the same protocol for automatic evaluation. Human evaluation is used to select the final winner. The latest results of automatic evaluation on LSMDC are publicly available~\cite{LSMDCresults}.
\vspace{-4mm}

\subsection{MSR-VTT}
In 2016, to further motivate and challenge the academic and the tech industry research community, Microsoft started the Microsoft Research - Video to Text (MSR-VTT)~\cite{msrvttchallengehome} competition aiming at bringing together computer vision and language researchers. The dataset used for this competition is MSR-VTT \cite{xu2016msr} described in the dataset section (Section~\ref{sec:datasets_general}). The participants of the competition are asked to develop a video to text model using MSR-VTT dataset. External datasets, either public or private can be used to help for better object, action, scene, and event detection, as long as the external data used are explicitly cited and explained in the submission file. 

\begin{figure*}[htbp] 
   \centering
   \includegraphics[width=0.8\textwidth]{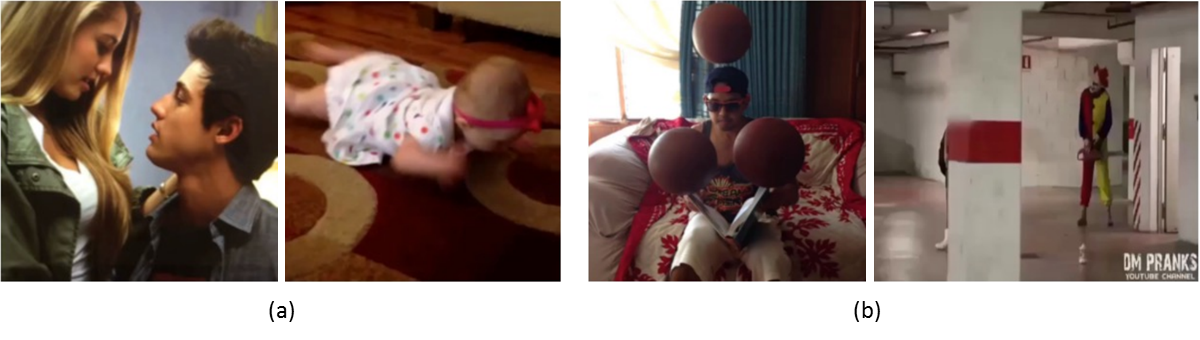} 
   \vspace{-4mm}
   \caption{Example video frames from TRECVID-VTT dataset. (a) Frames from the Easy-Video category and (b) frames from the Hard-Video category.}
   \label{fig:trecvidvttframes}
   \vspace{-6mm}
\end{figure*}

Unlike LSMDC, MSR-VTT challenge focuses only on the video to text task. This challenge requires a competing algorithm to automatically generate at least one natural language sentence that describes the most informative part in the video. Accuracy is benchmarked against human generated captions during the evaluation stage. The evaluation is based on an automatically computed score using multiple common metrics such as BLEU@4, METEOR, ROUGE-L, and CIDEr-D. Details of these metrics are given in Section-~\ref{sec:EvalMetPreamble}. Like LSMDC, human evaluations are also used to rank the generated sentences.
\vspace{-3mm}
\subsection{TRECVID}
Text Retrieval Conference (TREC) is a series of workshops emphasizing various subareas of Information Retrieval (IR) research. In particular, the TREC Video Retrieval Evaluation (TRECVID)~\cite{trecvidchallengehome} workshops, started in 2001, are dedicated to research efforts on content-based exploitation of digital videos. The primary areas of interests include ``semantic indexing, video summarization, video copy detection, multimedia event detection and ad-hoc video search''~\cite{trecvidchallengehome}. Since TREC-2016, Video to Text Description (VTT) \cite{awad2016trecvid} using natural language has also been included in the challenge tasks.

TRECVID-2017 VTT task used a dataset of over 50K automatically collected Twitter Vine videos, where each clip spans over approximately 6 seconds. This task is performed on a manually annotated selected subset that consists of 1,880 Twitter Vine videos. The dataset is further divided into four groups, G2, G3, G4 and G5, based on the number of descriptions (2 to 5) per videos. 
Furthermore, each video is tagged as easy or hard according to the difficulty level in describing it. Example frames from the VTT dataset are show in Figure~\ref{fig:trecvidvttframes}.

TRECVID uses metrics such as METEOR, BLEU and  CIDEr (details in Section-~\ref{sec:EvalMetPreamble}) for automatic evaluation, in addition to a newly introduced metric, referred to as Semantic Text Similarity (STS)~\cite{han2013umbc_ebiquity}. As the name suggests, STS measures semantic similarity of the generated and reference descriptions. Human evaluations are also employed to gauge the quality of the automatically generated descriptions following the Direct Assessment (DA) \cite{graham2017can} method. 
Due to its high reliability, DA is now employed as 
the official ranking method for machine translation benchmark evaluations~\cite{bojar2017findings}. As per DA based video description evaluation, human assessors are shown video-sentence pairs to rate how well the sentence describes the events in the video on a scale of ${0-100}$~ \cite{graham2017evaluation}. 

\subsection{ActivityNet Challenge}
ActivityNet Dense-Captioning Events in Videos ~\cite{activitynetcaptions} was first introduced in 2017 as a task of the ActivityNet Large Scale Activity Recognition Challenge~\cite{activitynethome, ghanem2017activitynet}, running as a CVPR Workshop since 2016. This task studies the detection and description of multiple events in a video. In the ActivityNet Captions Dataset, multiple descriptions along with time-stamps are provided for each video clip, where each description covers a unique portion of the clip.  Together, multiple events in that clip can be covered and narrated using the set of sentences. The events may be of variable durations (long or short) or even overlap. Details of this dataset are given in Section~\ref{subsec:activitynetcap} and Table~\ref{tab:datasets}.

Server based evaluations ~\cite{activitynetcaptionsevaluations} are performed for this challenge. The precision of captions generated are measured using BLEU, METEOR and CIDEr metrics. The latest results for the challenge are also publicly available and can be found online~\cite{activitynetcaptionsresults}. 


\section{Evaluation Metrics}
\label{sec:EvalMetPreamble}
Evaluations performed over machine generated captions/descriptions of videos can be divided into \textit{Automatic Evaluations} and \textit{Human Evaluations}. Automatic evaluations are performed using six different metrics which were originally designed for machine translation and image captioning. These metrics are BLEU~\cite{papineni2002bleu}, ROUGE$_L$~\cite{lin2004rouge}, METEOR~\cite{lavie2005meteor}, CIDEr~\cite{vedantam2015cider}, WMD~\cite{kusner2015word} and, SPICE~\cite{anderson2016spice}. Below, we discuss these metrics in detail as well as their limitations and reliability. Human Evaluations are performed to because of the unsatisfactory performance of automatic metrics given that there are numerous different ways to correctly describe the same video.

\subsection{Automatic Sentence Generation Evaluation}
\label{sec:evalmetricsintro}
\begin{figure*}[htbp] 
   \centering
   \includegraphics[width=\textwidth]{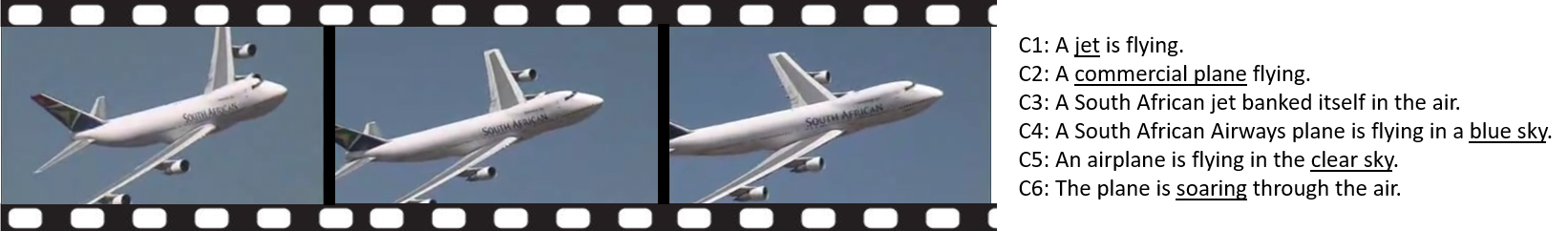} 
\vspace{-5mm}
   \caption{An example from MSVD~\cite{chen2011collecting} dataset with the associated ground truth captions. Note how the same video clip has been described very differently. Each caption describes the activity wholly or partially in a different way.}
\label{fig:plane}
\end{figure*}

Evaluation of video descriptions, automatically or manually generated, is challenging because as there is no specific ground truth or ``right answer'', that can be taken as a reference for benchmarking accuracy. A video can be correctly described in a wide variety of sentences, that may differ not only syntactically but also in terms of semantic content. Consider a sample from MSVD dataset as shown in Figure~\ref{fig:plane} for instance, several ground truth captions are available for the same video clip. Note that each caption describes the clip in an equally valid, but different way with varied attentions and levels of details in the clip, ranging from ``jet'', ``commercial airplane'' to ``South African jet'' and from ``flying'', ``soaring'' to ``banking'' and lastly from ``air'', ``blue sky'' to ``clear sky''. 

For automatic evaluation, when comparing the generated sentences with ground truth descriptions, three evaluation metrics are borrowed from machine translation, namely, Bilingual Evaluation Understudy (BLEU)~\cite{papineni2002bleu}, Recall Oriented Understudy of Gisting Evaluation (ROUGE)~\cite{lin2004rouge} and Metric for Evaluation of Translation with Explicit Ordering (METEOR)~\cite{lavie2005meteor}. Consensus based Image Description Evaluation (CIDEr)~\cite{vedantam2015cider} and Semantic Propositional Image Captioning Evaluation (SPICE)~\cite{anderson2016spice} are two other recently introduced metrics specifically designed for image captioning tasks, that are also being used for automatic evaluation of video description. 
Table~\ref{tab:evalmetrics} gives an overview of the metrics included in this survey. In addition to these automatic evaluation metrics, human evaluations are also employed to determine the performance of an automated video description algorithms.

\subsubsection{Bilingual Evaluation Understudy (BLEU, 2002)}
BLEU~\cite{papineni2002bleu} is a popular metric used to quantify the quality of machine generated text. The quality measures the correspondence between a machine and human outputs. BLEU scores take into account the overlap between predicted \textit{uni--gram}s (single word) or higher order \textit{n--gram} (sequence of $n$ adjacent words) and a set of one or more candidate reference sentences. According to BLEU, a high-scoring description should match the ground truth sentence in length i.e. exact match of words as well as their order. BLEU evaluation will score 1 for an exact match. Note that the more the number of reference sentences in the ground truth per video, the more the chances of a higher BLEU score. It is primarily designed to evaluate text at a corpus level and, therefore, its use as an evaluation metric over individual sentences may not be fair. BLEU is calculated as, 
\vspace{-3mm}
$$\log \text{BLEU} = \min (1-\frac{l_r}{l_c},0) + \sum_{n=1}^{N} w_n\log p_n.$$ 

\noindent
In the above equation, $l_r/l_c$ is the ratio between the lengths of the corresponding reference corpus and the candidate description, 
$w_n$ are positive weights, and $p_n$ is the geometric average of the modified n-gram precisions. While the second term computes the actual match score, the first term is a brevity penalty that penalizes descriptions that are shorter than the reference description. 

\subsubsection{Recall Oriented Understudy for Gisting Evaluation (ROUGE, 2004)}
ROUGE~\cite{lin2004rouge} metric was proposed in 2004 to evaluate text summaries. It calculates recall score of the generated sentences corresponding to the reference sentences using \textit{n--grams}. Similar to BLEU, ROUGE is also computed by varying the \textit{n--gram} count. However, unlike BLEU which is based on precision, ROUGE is based on recall values. Moreover, other than \textit{n--gram} variants of ROUGE$_n$, it has other versions known as , ROUGE$_L$ (Longest Common Subsequence), ROUGE$_W$ (Weighted Longest Common Subsequence), ROUGE$_S$ (Skip-Bigram Co-Occurrences Statistics), and ROUGE$_{SU}$ (extension of ROUGE$_S$). We refer the reader to the original paper for details. The version used in image and video captioning evaluation is ROUGE$_L$, which computes recall and precision scores of the longest common subsequences (LCS) between the generated and each reference sentence. The metric compares common subsequences of words in candidate and reference sentences. The intuition behind is that longer LCS of candidate and reference sentences corresponds to higher similarity between the two summaries. The words need not be consecutive but should be in sequence. ROUGE-N is computed as 

\vspace{-2mm}

$$\text{ROUGE-N}= \frac{\sum\limits_{S\in{R_{Sum}}}^{}\sum\limits_{g_n\in{S}}^{}C_{m}(g_n)}{\sum\limits_{S\in{R_{Sum}}}^{}\sum\limits_{g_n\in{S}}^{}C(g_n)},$$ \\
$n$ being the n-gram length, $g_n$, and $C_{m}(g_n)$ represents the highest number of n-grams that are present in candidate as well as ground truth summaries and R$_{Sum}$ stands for reference summaries. \\

LCS-based F-measure score is computed to find how similar summary $A$ of length $m$ is to summary $B$ of length $n$. Where $A$ is a sentence from the ground truth summary and $B$ is a sentence from the candidate generated summary. The recall $R_{lcs}$, precision $P_{lcs}$ and f-score $F_{lcs}$ are calculated as 
\vspace{-2mm}

$$R_{lcs} = \frac{\text{LCS}(A,B)}{m}, ~~~~~~ P_{lcs} = \frac{\text{LCS}(A,B)}{n},$$\\
\vspace{-3mm}
$$F_{lcs} = \frac{(1+\beta^2)R_{lcs}P_{lcs}}{R_{lcs} + \beta^2P_{lcs}},$$\\
where $\text{LCS}(A,B)$ is the length of longest common subsequence between A and B, $\beta = P_{lcs}/R_{lcs}$. The LCS-based F-measure score computed by equation $F_{lcs}$ is known as ROUGE$_L$ score. ROUGE$_L$ is 1 when $A=B$, and zero in case when A and B have no commonalities i.e. $\text{LCS}(A,B)=0$.

One of the advantages of ROUGE$_L$ is that it does not consider successive matches of words but employs in-sequence matches within a sentence. Moreover, pre-defining the n-gram length is also not required as this is automatically incorporated by $LCS$. 

\begin{table*}[htbp]
  \centering
  \small
   \setlength{\tabcolsep}{5.0pt}       
   \renewcommand{\arraystretch}{1.20}  
   \setlength{\belowcaptionskip}{-10pt}
   \caption{Summary of metrics used for video description evaluation.}
  \vspace{-4mm}
  \begin{tabular}{lll}
    \hline
    Metric Name & Designed For & Methodology \\
    \hline
    BLEU~\cite{papineni2002bleu}   & Machine translation    & \textit{n-gram} precision\\
    ROUGE~\cite{lin2004rouge}      & Document summarization & \textit{n-gram} recall \\
    METEOR~\cite{lavie2005meteor}  & Machine translation    & \textit{n-gram} with synonym matching \\
    CIDEr~\cite{vedantam2015cider} & Image captioning       & \textit{tf-idf} weighted \textit{n-gram} similarity \\
    SPICE~\cite{anderson2016spice} & Image captioning       & Scene-graph synonym matching\\
    WMD~\cite{kusner2015word}   & Document similarity    & Earth mover distance on word2vec\\
    \hline
    \end{tabular}%
  \label{tab:evalmetrics}%
\vspace{-3mm}
\end{table*}

\subsubsection{Metric for Evaluation of Translation with Explicit Ordering (METEOR, 2005)}
METEOR \cite{lavie2005meteor} was proposed to address the shortcomings of BLEU \cite{papineni2002bleu}. Instead of exact lexical match required by BLEU, METEOR introduced semantic matching. METEOR takes WordNet\cite{fellbaum1998wordnet}, a lexical database of the English language to account for various match levels, including exact words matches, stemmed words matches, synonymy matching and the paraphrase matching. 

METEOR score computation is based on how well the generated and reference sentences are aligned. Each sentence is taken as a set of unigrams and alignment is done by mapping unigrams of candidate and reference sentences. During mapping, a unigram in candidate sentence (or reference sentence) should either map to unigram in reference sentence (or candidate sentence) or to zero. In case of multiple options available for alignments between the two sentences, the alignment configuration with less number of crossings is preferred. After finalizing the alignment process, METEOR score is calculated.

Initially, unigram based precision score $P$ is calculated using $P=m_{cr}/m_{ct}$ relationship. 
Here $m_{cr}$ represents the number of unigrams co-occurring in both candidate, as well as reference sentences and $m_{ct}$ corresponds to total number of unigrams in the candidate sentences. Then unigram based recall score $R$ is calculated using $R=m_{cr}/m_{rt}$. Here $m_{cr}$ represents the number of unigrams co-occurring in both candidate as well as reference sentences. However, $m_{rt}$ is the number of unigrams in the reference sentences. Further, precision and recall scores are used to compute the F-score using following equation: 

\vspace{-2mm}
$$F_{mean}=\frac{10PR}{R+9P}.$$

The precision, recall and F-score measures account for unigram based congruity and do not cater for \textit{n--grams}. The \textit{n--gram} based similarities are used to calculate the penalty $p$ for alignment between candidate and reference sentences. This penalty takes into account the non-adjacent mappings between the two sentences. The penalty is calculated by grouping the unigrams into minimum number of chunks. The chunk includes unigrams that are adjacent in candidate as well as reference sentences. If a generated sentence is an exact match to the reference sentence then there will be only one chunk. The penalty is computed as 
$$p=\frac{1}{2}(\frac{N_c}{N_u})^2,$$ 
where $N_c$ in represents the number of chunks and $N_u$ corresponds to the number of unigrams grouped together. The METEOR score for the sentence is then computed as:
$$M=F_{mean} (1 - p).$$
Corpus level score can be computed using the same equation by using aggregated values of all the arguments i.e. $P, R$ and $p$. 
In case of multiple reference sentences, the maximum METEOR score of a generated and reference sentence is taken. To date, correlation of METEOR score with human judgments is better than that of BLEU score. Moreover, Elliot et al.~\cite{elliott2014comparing} also found METEOR to be a better evaluation metric as compared to contemporary metrics. Their conclusion is based on Spearman's correlation computation of automatic evaluation metrics against human judgments.

\subsubsection{Consensus based Image Description Evaluation (CIDEr, 2015)}
CIDEr \cite{vedantam2015cider} is a recently introduced evaluation metric for image
captioning task. It evaluates the consensus between a predicted sentence $c_i$ and reference sentences of the corresponding image. It performs stemming and converts all the words from candidate as well as reference sentences into their root forms e.g. \textit{stems, stemmer, stemming}, and \textit{stemmed} to their root word \textit{stem}. CIDEr treats each sentence as a set of \textit{n--grams} containing 1 to 4 words. To encode the consensus between predicted sentence and reference sentence, it measures the co-existence frequency of n-grams in both sentences. Finally, \textit{n--grams} that are very common among the reference sentences of all the images are given lower weight, as they are likely to be less informative about the image content, and more biased towards lexical structure of the sentences. The weight for each \textit{n--gram} is computed using Term Frequency Inverse Document Frequency (TF-IDF)~\cite{robertson2004understanding}. The term TF puts higher weightage on frequently occurring \textit{n--grams} in the reference sentence of the image, whereas IDF puts lower weightage on commonly appearing \textit{n--grams} across the whole dataset.

Finally, CIDEr$_n$ score is computed as 

$$\text{CIDEr}_n(c_i, S_i) = \frac{1}{m}\sum_{j}^{}\frac{g^{n}(c_i).g^{n}(s_{ij})}{\norm{g^{n}(c_i)}.\norm{g^{n}(s_{ij})}},$$
where $g^{n}(c_i)$ is a vector representing all \textit{n--grams} with length $n$ and $\norm{g^{n}(c_i)}$  depicts magnitude of $g^{n}(c_i)$. Same is true for $g^{n}(s_{ij})$. Further, CIDEr uses higher order n-grams (higher the order, longer the sequence of words) to capture the grammatical properties and richer semantics of the text. For that matter, it combines the scores of different n-grams using the following equation:

$$\text{CIDEr}(c_i, S_i) = \sum_{n=1}^{N}w_n \text{CIDEr}_n(c_i, S_i).$$

The most popular version of CIDEr in image and video description evaluation is CIDEr-D, that incorporates a few modifications in the originally proposed CIDEr to prevent higher scores for the captions that badly fail in human judgments. Firstly, they proposed removal of stemming to ensure correct form of words are used. Otherwise, multiple forms of verbs (singular, plural etc) are mapped to the same token producing high score for incorrect sentences. Secondly, they ensure that if the words of high confidence are repeated in a sentence a high score is not produced as in the original CIDEr produces even if the sentence does not make sense. This is done by introducing a Gaussian penalty over length differences between the candidate and reference sentences and by clipping to the \textit{n--grams} count equal to the number of occurrences in the reference sentence. The latter ensures that the desired sentence length is not achieved by repetition of high confidence words to get a high score. The aforementioned changes makes the metric robust and ensures its high correlation score~\cite{vedantam2015cider}.

\begin{figure}[b] 
   \centering
   \includegraphics[width=0.9\columnwidth]{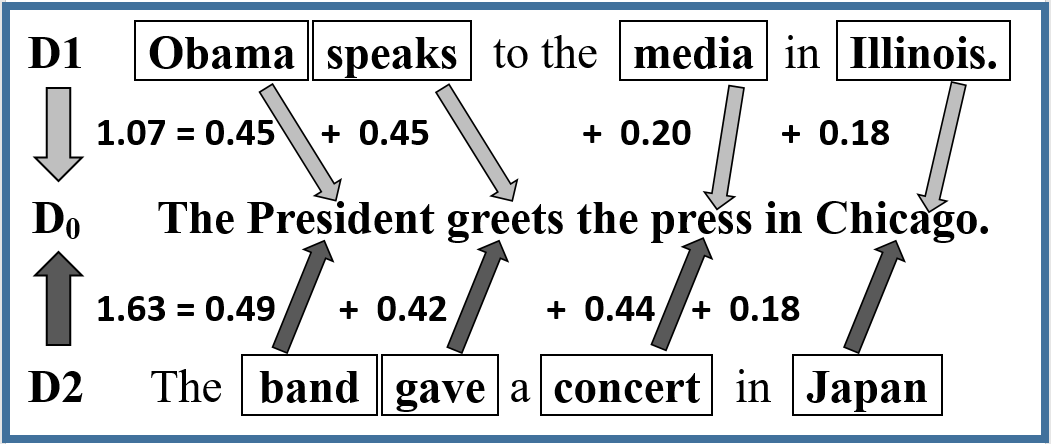} 
\vspace{-2mm}
   \caption{Components of the WMD metric between a query D$_0$ and two sentences $D_1$ and $D_2$ with the same BOW distance. $D_1$ with less distance 1.07 matches with query D$_0$ than $D_2$ with distance 1.63. The arrows show flow between two words and are labeled with their distance contribution. Figure adapted from~\cite{kusner2015word}.}
   \label{fig:wmdmetric}
\end{figure}

\begin{table*}[ht]
  \small
  \centering
  \caption{Variations in automatic evaluation metric scores with four types of changes made to candidate sentence i.e. words replaced with their synonyms, added redundancy to sentence, changing word order, and shortening the sentence length. The first row shows the upper bound scores of BLEU-4, METEOR, ROUGE, and CIDEr represented by B, M, R, and C respectively.}
  \vspace{-4mm}
    \begin{tabular}{lp{20 em}rrrr}
    \hline
    Variation & Description & B & M & R & C \\
    \hline
    reference & an elderly man is playing piano in front of a crowd in an anteroom & 1 & 1 & 1 & 10 \\
    candidate  & an elderly man is showing how to play piano in front of a crowd in a hall room & 0.47 & 0.45 & 0.70 & 0.53 \\
    \hline
    synonyms & an \textbf{old} man is \textbf{demonstrating} how to play piano in front of a crowd in a hall room & 0.37 & 0.40 & 0.64 & 0.43 \\
    redundancy & an elderly man is showing how to play piano in front of a crowd in a hall room \textbf{with a woman} & 0.40 & 0.44  & 0.65 & 0.47 \\
    word order &  an elderly man in front of a crowd is showing how to play piano in a hall room & 0.30 & 0.39 & 0.57 & 0.35 \\
    short length &  a man is playing piano & 0.12 & 0.22 & 0.39 & 0.49 \\
    \hline
    \end{tabular}%
  \label{tab:evalmetricscompare}%
\end{table*}

\subsubsection{Word Mover's Distance (WMD, 2015)}
The WMD~\cite{kusner2015word} makes use of word embeddings which are semantically meaningful vector representations of words learnt from text corpora. WMD distance measures the dissimilarity between two text documents. Two captions with different words may still have the same semantic meanings. On the other hand, it is possible for multiple captions to have the same attributes, objects and their relations while still having very different meanings. WMD was proposed to address this problem. This is because word embeddings are good at capturing semantic meanings and are easier to compute than WordNet thanks to the distributed vector representations of words. The distance between two texts is casted as an Earth Mover's Distance (EMD) \cite{rubner2000earth}, typically used in transportation to calculate the travel cost using word2vec embeddings \cite{mikolov2013distributed}.

In this metric, each caption or description is represented by a bag-of-words histogram that includes all but the start and stop words. The magnitude of each bag-of-words histogram is then normalized.
To account for semantic similarities that exist between pairs of words, the WMD metric uses the Euclidean distance in the word2vec embedding space. The distance between two documents or captions is then defined as the cost required to move all words between captions. Figure~\ref{fig:wmdmetric} illustrates an example WMD calculation process. 
The WMD is modelled as a special case of EMD~\cite{rubner2000earth} and is then solved by linear optimization. Compared to BLUE, ROUGE and CIDEr, WMD is less sensitive to words order or synonym swapping. Further, similar to CIDEr and METEOR, it gives high correlation against human judgments.

\subsubsection{Semantic Propositional Image Captioning Evaluation (SPICE, 2016)}
SPICE \cite{anderson2016spice} is the latest proposed evaluation metric for image and video descriptions. SPICE measures the similarity between the {\em scene graph tuples} parsed from the machine generated descriptions and the ground truth. The semantic scene graph encodes objects, their attributes and relationships through a dependency parse tree. A scene graph tuple $G(c)$ of caption $c$ consists of semantic tokens such as object classes $O(c)$, relation types $R(c)$ and attribute types $A(c)$,
$$G(c)=\langle O(c),R(c),A(c) \rangle .$$
SPICE is computed based on F1-score between the tuples of machine generated descriptions and the ground truth. Like METEOR, SPICE also uses WordNet to find and treat synonyms as positive matches. Although, in the current literature, the SPICE score has not been employed much but one obvious limiting factor on its performance could be the quality of the parsing. For instance, in a sentence {\tt ``white dog swimming through river}'', the failure case could be the word ``\textit{swimming}'' being parsed as ``\textit{object}'' and the word ``\textit{dog}'' parsed as ``\textit{attribute}'' resulting in a very bad score.

\subsection{Human Evaluations}
Given the lack of reference captions and low correlation with human judgments of automated evaluation metrics, human evaluations are also often used to judge the quality of machine generated captions. Human evaluations may either be crowd-sourced, such as AMT workers or specialist judges as in some competitions. Such human evaluations can be  further structured using measurements such as \textit{Relevance} or \textit{Grammar Correctness}. In relevance based evaluation, video content relevance is given subjective scores, with highest score given to the \textit{``Most Relevant''} and minimum score to the \textit{``Least Relevant''}. The score of two sentences cannot be the same unless they are identical. In the approaches where grammar correctness is measured, the sentences are graded based on grammatical correctness without showing the video content to the evaluators in which case, more than one sentence may have the same score.

\vspace{-3mm}
\subsection{Limitations of Evaluation Metrics}
Like video description, evaluation of the machine generated sentences is an equally difficult task. There is no metric specifically designed for evaluating video description, instead machine translation and image captioning metrics have been extended for this task. These automatic metrics compute the score given reference and candidate sentences. This paradigm has a serious problem that there can be several different ways to describe the same video, all correct at the same time, depending upon \textit{``what has been described}'' (content selection) and \textit{``how it has been described''} (realization). These metrics fail to incorporate all these variations and are, therefore, far from being perfect. 
Various studies~\cite{kilickaya2016re, wang2016cross} have examined how metric scores behave under different conditions. In Table~\ref{tab:evalmetricscompare}, we perform similar experiments~\cite{kilickaya2016re} but with an additional variation of \emph{short length}. First, the original caption was evaluated with itself to analyze the maximum possible score achievable by each metric (first row of Table \ref{tab:evalmetricscompare}). Next, minor modifications were introduced in the candidate sentences to measure how the evaluation metrics behave. It was observed that all metric scores reduced, BLEU and CIDEr being the most affected, when some words were replaced with their synonyms. This is apparently due to the failure to match synonyms. Further experiments revealed that the metrics were generally stable when the sentence was perturbed with a few additional words. However, changing the word order in a sentence was found to alter the scores of {\em n-gram} based metrics like BLEU, ROUGE and CIDEr significantly and that of ROUGE to some extent. On the other hand, WMD and SPICE were found to be robust to word order changes~\cite{kilickaya2016re}. Lastly, reducing the sentence length significantly affected BLEU, METEOR and ROUGE scores but had little effect on CIDEr score i.e. the scores were reduced by 74\%, 51\%, 44\% and 7\% respectively.

\subsection{Reliability of Evaluation Metrics}
A good method to evaluate the video descriptions is to compare the machine generated descriptions with the ground truth descriptions annotated by humans. However, as shown in Figure~\ref{fig:plane}, the reference captions can vary within itself and can only represent few samples out of all valid samples for the same video clip. Having more reference sample captions create a better solution space and hence lead to more reliable evaluation.

Another aspect of the evaluation problem is the syntactic variations in candidate sentences. The same problem also exists in the well studied field of machine translation. In this case, a sentence in a source language can be translated into various sentences in a target language. Syntactically different sentences may still have the same semantic content.

In a nutshell, evaluation metrics assess the suitability of a caption to the visual input by comparing how well the candidate caption matches with that of reference caption(s). The agreement of the metric scores with human judgments (i.e.~the gold standard) improves with the increased number of reference captions \cite{vedantam2015cider}. 
Numerous studies \cite{vedantam2015cider, vedantam2015cider, pan2016hierarchical, yu2016video, venugopalan2015sequence} also found that CIDEr, WMD, SPICE and METEOR have higher correlations to human judgments and are regarded as superior amongst the contemporary metrics. WMD and SPICE are very recent automatic caption evaluation metrics and have not been studied extensively in the literature at the time of this survey.
\begin{table*}[htbp]
  \centering
  \small
  \setlength{\tabcolsep}{10.0pt}       
  \caption{Performance of video captioning methods on MSVD dataset. Higher scores are better in all metrics. The best score for each metric is shown in bold. }
  \vspace{-2mm}
  \begin{tabular}{|p{15.43em}|c|c|c|c|c|c|}
    \hline
    \multirow{2}[1]{*}{\textbf{Techniques / Models / Methods}} & \multirow{2}[1]{*}{\textbf{Yr}} & \multirow{2}[1]{*}{\textbf{Dataset}} & \multicolumn{4}{c|}{\textbf{Results}} \\
  \cline{4-7}    \multicolumn{1}{|l|}{} & & & \textbf{BLEU} & \textbf{METEOR} & \textbf{CIDEr} & \textbf{ROUGE} \\
    \hline
    RBS+RBS \& RF-TP+RBS \cite{hanckmann2012automated} & 2012  & MSVD  & \multicolumn{4}{c|}{SVO Accuracy} \\
    \hline 
    SVO-LM (VE)~\cite{krishnamoorthy2013generating} & 2013  & MSVD  & 0.45+\_0.05 & 0.36+\_0.27 &       &  \\
    \hline
    FGM~\cite{thomason2014integrating}  & 2014  & MSVD  & \multicolumn{4}{c|}{SVOP Accuracy} \\
    \hline 
    LSTM-YT~\cite{venugopalan2014translating} & 2015  & {MSVD} & 33.3  & 29.1  & -     & - \\
    \hline
    TA~\cite{yao2015describing} & 2015  & MSVD  & 41.9  & 29.6  & 51.67  & - \\
    \hline
    S2VT~\cite{venugopalan2015sequence} & 2015  & {MSVD} & -     & 29.8  & -     & - \\
    \hline
    \multicolumn{1}{|l|}{h-RNN~\cite{yu2016video}}  & 2016  & {MSVD} & 49.9  & 32.6  & 65.8  & - \\
    \hline 
    \multicolumn{1}{|l|}{MM-VDN~\cite{xu2015multi}}  & 2016  & MSVD  & 37.6  & 29.0  & -     & - \\
    \hline
     \multicolumn{1}{|l|}{Glove + Deep Fusion Ensble~\cite{venugopalan2016improving}}  & 2016  & MSVD  & 42.1 & 31.4  & -     & - \\
    \hline
     \multicolumn{1}{|l|}{S2FT~\cite{liu2016boosting}}  & 2016  & MSVD  & -  & 29.9  & - & - \\
    \hline
    \multicolumn{1}{|l|}{HRNE~\cite{pan2016hierarchical}}  & 2016  & MSVD  & 43.8  & 33.1  & -     & - \\
    \hline
    \multicolumn{1}{|l|}{GRU-RCN~\cite{ballas2015delving}}  & 2016  & MSVD  & 43.3  & 31.6  & 68.0 & - \\
    \hline
    LSTM-E~\cite{pan2016jointly} & 2016  & MSVD  & 45.3  & 31.0  & - & - \\
    \hline
    \multicolumn{1}{|l|}{SCN-LSTM~\cite{gan2017semantic}} & 2017  & MSVD  & 51.1  & 33.5 & \textbf{77.7}  & - \\
    \hline
    \multicolumn{1}{|l|}{LSTM-TSA~\cite{Pan_2017_CVPR}} & 2017  & MSVD  & \textbf{52.8}  & 33.5 & 74.0  & - \\
    \hline
    \multicolumn{1}{|l|}{TDDF~\cite{Zhang_2017_CVPR}} & 2017  & MSVD  & 45.8  & 33.3  & 73.0  & 69.7 \\
    \hline
    \multicolumn{1}{|l|}{BAE~\cite{Baraldi_2017_CVPR}} & 2017  & MSVD  & 42.5  & 32.4 & 63.5  & - \\
    \hline
    \multicolumn{1}{|l|}{PickNet~\cite{chen2018less}} & 2018  & MSVD  & 46.1 & 33.1 & 76.0  & 69.2 \\
    \hline
    \multicolumn{1}{|l|}{M$^3-{IC}$~\cite{wang2018m3}} & 2018  & MSVD  & \textbf{52.8} & 33.3 & -  & - \\
    \hline
    RecNet$_{local}$~\cite{wang2018reconstruction} & 2018  & MSVD & 52.3 & 34.1 & \textbf{80.3} & 69.8 \\
    \hline
    TSA-ED~\cite{wu2018interpretable} & 2018  & MSVD & 51.7 & 34.0 & 74.9 & - \\
    \hline
    GRU-EVE~\cite{gruevehftsem} & 2019 & MSVD & 47.9 & \textbf{35.0} & 78.1 & \textbf{71.5} \\
    \hline

    \end{tabular}%
  \label{tab:msvdresults}%
 \end{table*}%

\vspace{-3mm}
\section{Benchmark Results}
\label{ch:results}
We summarize the benchmark results of various techniques on each video description dataset. We group the methods based on the dataset they reported results on and then order them chronologically. Moreover, for multiple variants of the same model, only their best reported results are reported here. For a detailed analysis of each method and its variants, the original paper should be consulted. In addition, where multiple \textit{n--gram} scores are reported for the BLEU metric, we have chosen only the BLEU@4 results as this is the closest to human evaluations. From Table~\ref{tab:msvdresults}, we can see that most methods have reported results on the MSVD dataset, followed by MSR-VTT, M-VAD, MPII-MD, and ActivityNet Captions. The popularity of MSVD can be attributed to the diverse nature of YouTube videos and the large number of reference captioning. MPII-MD, M-VAD, MSR-VTT and ActivityNet Captions are popular because of their size and their inclusion in competitions (see Section~\ref{videocompetitions}).

\begin{table*}[htbp]
  \centering
  \small
  \setlength{\tabcolsep}{10.0pt}       

\caption{Performance of video captioning methods on TACoS-MLevel dataset. Higher scores are better in all metrics. The best score for each metric is shown in bold.}
\vspace{-2mm}
     \begin{tabular}{|p{15.43em}|c|c|c|c|c|c|}
    \hline
    \multirow{2}[1]{*}{\textbf{Techniques / Models / Methods}} & \multirow{2}[1]{*}{\textbf{Yr}} & \multirow{2}[1]{*}{\textbf{Dataset}} & \multicolumn{4}{c|}{\textbf{Results}} \\
\cline{4-7}    \multicolumn{1}{|l|}{} & & & \textbf{BLEU} & \textbf{METEOR} & \textbf{CIDEr} & \textbf{ROUGE} \\
    \hline
    SMT(SR) + Prob I/P \cite{rohrbach2014coherent} & 2014  & TACoS MLevel & 28.5  & -     & -     & - \\
    \hline
    \multicolumn{1}{|l|}{CRF + LSTM-Decoder \cite{donahue2015long}}  & 2015  & {TACoS MLevel} & 28.8  & -     & -     & - \\
    \hline
    \multicolumn{1}{|l|}{h-RNN \cite{yu2016video}}  & 2016  & TACoS MLevel & \textbf{30.5}  & \textbf{28.7}  & \textbf{160.2}  & - \\
    \hline
    \multicolumn{1}{|l|}{JEDDi-Net \cite{xu2018joint}} & 2018 & TACoS MLevel & 18.1 & 23.85 & 103.98 & 50.85 \\
    \hline
    \end{tabular}%
  \label{tab:tacosmlevelresults}%
 \end{table*}%

\begin{table*}[htbp]
  \centering
  \small
  \setlength{\tabcolsep}{10.0pt}       
\caption{Performance of video captioning methods on M-VAD dataset.}
\vspace{-2mm}
     \begin{tabular}{|p{15.43em}|c|c|c|c|c|c|}
    \hline
    \multirow{2}[1]{*}{\textbf{Techniques / Models / Methods}} & \multirow{2}[1]{*}{\textbf{Yr}} & \multirow{2}[1]{*}{\textbf{Dataset}} & \multicolumn{4}{c|}{\textbf{Results}} \\
\cline{4-7}    \multicolumn{1}{|l|}{} & & & \textbf{BLEU} & \textbf{METEOR} & \textbf{CIDEr} & \textbf{ROUGE} \\
    \hline
    Temporal-Attention (TA) \cite{yao2015describing} & 2015  & {M-VAD} & \textbf{0.7}   & 5.7   & 6.1   & - \\
    \hline
    S2VT \cite{venugopalan2015sequence} & 2015  & {M-VAD} & -     & 6.7   & -     & - \\
    \hline
    Visual-Labels \cite{rohrbach2015long} & 2015  & {M-VAD} & - & 6.4 & - & - \\
    \hline
    \multicolumn{1}{|l|}{HRNE \cite{pan2016hierarchical}}  & 2016  & {M-VAD} & \textbf{0.7}   & 6.8   & -     & - \\
    \hline
    \multicolumn{1}{|l|}{Glove + Deep Fusion Ensemble \cite{venugopalan2016improving}}  & 2016  & M-VAD  & - & 6.8  & - & - \\
    \hline
    LSTM-E \cite{pan2016jointly} & 2016  & {M-VAD} & -     & 6.7   & -     & - \\
    \hline
    \multicolumn{1}{|l|}{LSTM-TSA \cite{Pan_2017_CVPR}} & 2017  & M-VAD  & -  & 7.2 & - & - \\
    \hline
    \multicolumn{1}{|l|}{BAE \cite{Baraldi_2017_CVPR}} & 2017  & M-VAD & -  & \textbf{7.3} & -  & - \\
    \hline
    \end{tabular}%
  \label{tab:mvadresults}%
 \end{table*}%

\begin{table*}[htbp]
  \centering
  \small 
  \setlength{\tabcolsep}{10.0pt}       
\caption{Performance of video captioning methods on MPII-MD dataset.}
 \vspace{-2mm}
    \begin{tabular}{|p{15.43em}|c|c|c|c|c|c|}
    \hline
    \multirow{2}[1]{*}{\textbf{Techniques / Models / Methods}} & \multirow{2}[1]{*}{\textbf{Yr}} & \multirow{2}[1]{*}{\textbf{Dataset}} & \multicolumn{4}{c|}{\textbf{Results}} \\
\cline{4-7}    \multicolumn{1}{|l|}{} & & & \textbf{BLEU} & \textbf{METEOR} & \textbf{CIDEr} & \textbf{ROUGE} \\
    \hline

    S2VT \cite{venugopalan2015sequence} & 2015  & {MPII-MD} & -     & 7.1   & -     & - \\
    \hline
    Visual-Labels \cite{rohrbach2015long} & 2015 & {MPII-MD} & - & 7.0 & - & - \\
    \hline
    SMT \cite{rohrbach2015dataset} & 2015 & {MPII-MD} & - & 5.6 & - & - \\
    \hline
    \multicolumn{1}{|l|}{Glove + Deep Fusion Ensemble \cite{venugopalan2016improving}}  & 2016  & MPII-MD  & - & 6.8  & - & - \\
    \hline
    LSTM-E \cite{pan2016jointly} & 2016  & MPII-MD & -     & 7.3   & -     & - \\
    \hline
    \multicolumn{1}{|l|}{LSTM-TSA \cite{Pan_2017_CVPR}} & 2017  & MPII-MD  & -  & \textbf{8.0} & - & - \\
    \hline
    \multicolumn{1}{|l|}{BAE \cite{Baraldi_2017_CVPR}} & 2017  & MPII-MD & \textbf{0.8} & 7.0 & \textbf{10.8} & \textbf{16.7} \\
    \hline

    \end{tabular}%
  \label{tab:mpiimdresults}%
 \end{table*}%
\begin{table*}[bhtp]
  \centering
  \small
  \setlength{\tabcolsep}{10.0pt}       
\caption{Performance of video captioning methods on MSR-VTT dataset..}
\vspace{-2mm}
     \begin{tabular}{|p{15.43em}|c|c|c|c|c|c|}
    \hline
    \multirow{2}[1]{*}{\textbf{Techniques / Models / Methods}} & \multirow{2}[1]{*}{\textbf{Yr}} & \multirow{2}[1]{*}{\textbf{Dataset}} & \multicolumn{4}{c|}{\textbf{Results}} \\
\cline{4-7}    \multicolumn{1}{|l|}{} & & & \textbf{BLEU} & \textbf{METEOR} & \textbf{CIDEr} & \textbf{ROUGE} \\
    \hline
    \multicolumn{1}{|l|}{Alto \cite{shetty2016frame}} & 2016  & MSR-VTT  & 39.8  & 26.9 & 45.7 & 59.8 \\
    \hline
     \multicolumn{1}{|l|}{VideoLab \cite{ramanishka2016multimodal}} & 2016  & MSR-VTT  & 39.1  & 27.7 & 44.4 & 60.6 \\
    \hline
    \multicolumn{1}{|l|}{RUC-UVA \cite{dong2016early}} & 2016  & MSR-VTT  & 38.7 & 26.9 & 45.9 & 58.7 \\
    \hline
    \multicolumn{1}{|l|}{v2t-navigator \cite{jin2016describing}} & 2016  & MSR-VTT  & 40.8  & 28.2 & 44.8 & 61.1 \\
    \hline
    \multicolumn{1}{|l|}{TDDF \cite{Zhang_2017_CVPR}} & 2017  & MSR-VTT  & 37.3  & 27.8 & 43.8 & 59.2 \\
    \hline
    \multicolumn{1}{|l|}{DenseVidCap \cite{Shen_2017_CVPR}} & 2017  & MSR-VTT  & 41.4 & 28.3 & 48.9 & 61.1 \\
    \hline
     \multicolumn{1}{|l|}{CST-GT-None \cite{phan2017consensus}} & 2017  & MSR-VTT  & \textbf{44.1} & \textbf{29.1} & \textbf{49.7} & \textbf{62.4} \\
    \hline
    \multicolumn{1}{|l|}{PickNet \cite{chen2018less}} & 2018  & MSR-VTT & 38.9 & 27.2 & 42.1 & 59.5 \\
    \hline
    HRL \cite{wang2017video} & 2018  & MSR-VTT & 41.3  & 28.7  & 48.0  & 61.7 \\
    \hline
    \multicolumn{1}{|l|}{M$^3-{VC}$~\cite{wang2018m3}} & 2018  & MSR-VTT  & 38.1 & 26.6 & -  & - \\
    \hline
    RecNet$_{local}$~\cite{wang2018reconstruction} & 2018  & MSR-VTT & 39.1 & 26.6 & 42.7 & 59.3 \\
    \hline
    GRU-EVE~\cite{gruevehftsem} & 2019 & MSR-VTT & 38.3 & 28.4 & 48.1 & 60.7 \\
    \hline
    \end{tabular}%
  \label{tab:msrvttresults}%
 \end{table*}%
Another key observation is that earlier works have mainly reported results in terms of subject, verb, object (SVO) and in some cases place (scene) detection accuracies in the video, whereas more recent works started to report sentence level matches using the automatic evaluation metrics. Considering the diverse nature of the datasets and the limitations of automatic evaluation metrics, we analyze the results of different methods using four popular metrics namely BLEU, METEOR, CIDEr and ROUGE. 

\begin{table*}[htbp]
  \centering
  \small
  \setlength{\tabcolsep}{10.0pt}       

\caption{Performance of video captioning methods on ActivityNet Captions dataset.}
\vspace{-2mm}
     \begin{tabular}{|p{15.43em}|c|c|c|c|c|c|}
    \hline
    \multirow{2}[1]{*}{\textbf{Techniques / Models / Methods}} & \multirow{2}[1]{*}{\textbf{Yr}} & \multirow{2}[1]{*}{\textbf{Dataset}} & \multicolumn{4}{c|}{\textbf{Results}} \\
\cline{4-7}    \multicolumn{1}{|l|}{} & & & \textbf{BLEU} & \textbf{METEOR} & \textbf{CIDEr} & \textbf{ROUGE} \\
    \hline
     Dense-Cap Model~\cite{krishna2017dense} & 2017  & ActivityNet Cap & \textbf{3.98}  & 9.5   & 24.6  & - \\
    \hline
    LSTM-A+PG+R~\cite{yaomsr} & 2017  & ActivityNet Cap & -  & \textbf{12.84} & - & - \\
    \hline
     TAC \cite{ruccmu} & 2017  & ActivityNet Cap & - & 9.61 & - & - \\
    \hline
    JEDDi-Net~\cite{xu2018joint} & 2018 & ActivityNet Cap & 1.63 & 8.58 & 19.88 & \textbf{19.63} \\
    \hline 
    DVC~\cite{li2018jointly} & 2018 & ActivityNet Cap & 1.62 & 10.33 & \textbf{25.24} & - \\
    \hline
    Bi-SST~\cite{wang2018bidirectional} & 2018 & ActivityNet Cap & 2.30 & 9.60 & 12.68 & 19.10 \\
    \hline
    Masked Transformer~\cite{zhou2018end} & 2018 & ActivityNet Cap & 2.23 & 9.56 & - & - \\
    \hline
    \end{tabular}%
  \label{tab:activitynetcapresults}%
 \end{table*}%
\begin{table*}[htbp]
  \centering
  \small
  \setlength{\tabcolsep}{10.0pt}       

\caption{Performance of video captioning methods on various benchmark datasets.}
 \vspace{-2mm}
    \begin{tabular}{|p{15.43em}|c|c|c|c|c|c|}
    \hline
    \multirow{2}[1]{*}{\textbf{Techniques / Models / Methods}} & \multirow{2}[1]{*}{\textbf{Yr}} & \multirow{2}[1]{*}{\textbf{Dataset}} & \multicolumn{4}{c|}{\textbf{Results}} \\
\cline{4-7}    \multicolumn{1}{|l|}{} & & & \textbf{BLEU} & \textbf{METEOR} & \textbf{CIDEr} & \textbf{ROUGE} \\
    \hline

    CT-SAN \cite{yu2016end}  & 2016  & LSMDC & \textbf{0.8}  & 7.1  & \textbf{10.0}  & \textbf{15.9} \\
    \hline
    GEAN \cite{Yu_2017_CVPR} & 2017  & LSMDC & - & \textbf{7.2} & 9.3 & 15.6 \\
    \hline
    \hline
    HRL \cite{wang2017video}  & 2018  & Charades & \textbf{18.8}  & \textbf{19.5}  & \textbf{23.2}  & \textbf{41.4} \\
    \hline
    TSA-ED~\cite{wu2018interpretable} & 2018  & Charades & 13.5 & 17.8 & 20.8 & - \\
    \hline
    \hline
    Masked Transformer~\cite{zhou2018end} & 2018 & YouCook-II & \textbf{1.13} & \textbf{5.90} & - & - \\
    \hline
    \end{tabular}%
  \label{tab:miscdatasetsresults}%
 \end{table*}%

Table~\ref{tab:msvdresults} summarizes results for the MSVD dataset. GRU-EVE~\cite{gruevehftsem} 
achieves the best performance on METEOR and ROUGE$_L$ metrics and the second best on CIDEr metric whereas LSTM-TSA \cite{Pan_2017_CVPR} and M$^3$-${IC}$~\cite{wang2018m3} report the best BLEU scores. RecNet$_{local}$~\cite{wang2018reconstruction} has the best CIDEr score and second best BLEU score. As shown in Table~\ref{tab:tacosmlevelresults}, on TACoS Multilevel dataset, h-RNN \cite{yu2016video} has the best results on all reported metrics i.e. BLEU, METEOR and CIDEr. This method does not provide ROUGE score.

On the more challenging M-VAD dataset, overall the reported results (Table~\ref{tab:mvadresults}) are very poor, however, within the presented results we see that so far only Temporal-Attention \cite{yao2015describing}, and HRNE \cite{pan2016hierarchical} reported results using the BLEU metric with a BLEU score of 0.7 each. All the papers using this dataset report METEOR results and so far BAE \cite{Baraldi_2017_CVPR} has produced the best METEOR score followed by LSTM-TSA \cite{Pan_2017_CVPR}. HRNE \cite{pan2016hierarchical} and Glove+Deep Fusion Ensemble \cite{venugopalan2016improving} share the third place for METEOR score.

MPII-MD is another very challenging dataset and still has very low benchmark results, as shown in Table~\ref{tab:mpiimdresults}, similar to the M-VAD dataset. Only BAE~\cite{Baraldi_2017_CVPR} has reported BLEU score for this dataset. LSTM-TSA \cite{Pan_2017_CVPR} has achieved the best METEOR score followed by LSTM-E \cite{pan2016jointly} and S2VT \cite{venugopalan2015sequence} at second and third place respectively. No other paper using this dataset has reported CIDEr and ROUGE score except BAE~\cite{Baraldi_2017_CVPR}. 

Results on another popular dataset, MSR-VTT, are overall better than the M-VAD and MPII-II datasets. As shown in Table~\ref{tab:msrvttresults}, CST-GT-None~\cite{phan2017consensus} has reported the highest score on all four metrics i.e. BLEU, METEOR, CIDEr and ROUGE. 
DenseVidCap~\cite{Shen_2017_CVPR} and HRL~\cite{wang2017video} respectively report the second and third best scores on BLEU metric. GRU-EVE~\cite{gruevehftsem} reports the third best score in METEOR and CIDEr metrics.

Results of another recent and popular ActivityNet Captions dataset are presented in Table~\ref{tab:activitynetcapresults}. This dataset was primarily introduced for dense video captioning and is gaining popularity very quickly. In this dataset, Dense-Cap Model~\cite{krishna2017dense} stands at top in terms of BLEU score. Best METEOR score is reported by LSTM-A+PG+R~\cite{yaomsr}. Highest scores in CIDEr and ROUGE metrics are achieved by methods DVC~\cite{li2018jointly} and JEDDi-Net~\cite{xu2018joint} respectively.
Finally, in Table~\ref{tab:miscdatasetsresults}, we report two results for LSMDC and Charades each and only one result for YouCook-II datasets. YouCook-II is also a recent dataset and not reported much in the literature.

We summarize the best reporting methods for each dataset along with their published scores. The tables group methods by the used dataset(s). Hence, one can infer the difficulty level of datasets by comparing the intra dataset scores of the same methods and the popularity of a particular dataset from the number of methods that have reported results on it.


\section{Future and Emerging Directions}
\label{sec:futuredirections}
Automatic video description has come very far since the pioneer methods, especially after the adoption of deep learning. Although the performance of existing methods is still far below that of humans, the gap is diminishing at a steady rate and there is still ample room for algorithmic improvements. Here, we list several possible future and emerging directions that have the potential to advance this research area.

\vspace{2mm}
\noindent {\bf Visual Reasoning:}
Although video VQA is still in nascent stage, beyond VQA is the visual reasoning problem. This is a very promising field to further explore. Here the model is made not to just answer a particular question but to reason why it chose that particular answer.  For example in a video where a road side with parking marks is shown, the question is ``\textit{Can a vehicle be parked here?}'', the model answers correctly, ``\textit{Yes}''. The next question is ``\textit{Why?}'' to which the model reasons that there is a parking sign on the road which means it is legal to park here. Another example is the explanations generated by self driving cars \cite{explanable_selfDriving} where the system keeps the passengers in confidence by generating natural language descriptions of the reasons behind its decisions e.g. to slow down, take a turn etc. An example of visual reasoning models is the MAC Network~\cite{macnetreason} which is able to think and reason giving promising results on CLEVR~\cite{cleverreasondataset}, a visual reasoning dataset.


\vspace{1mm}
\noindent {\bf Visual Dialogue:}
Similar to audio dialogue (e.g.~Siri, Hello Google, Alexa and ECHO), visual dialogue~\cite{visualdialogue} is another promising and flourishing field, especially in an era where we look forward to interact with robots. In visual dialogue, given a video, a model is asked a series of questions sequentially in a dialogue/conversation manner. The model tries to answer (no matter right or wrong) these questions. This is different from visual reasoning where the model argues the reasons that lead the model to choose particular answers.

\vspace{1mm}
\noindent {\bf Audio and Video:}
While the majority of computer vision research has focused on video description, without the help of audio, audio is naturally present in most of videos. Audio can help in video description by providing background information for instance, the sound of train, ocean, traffic when there is no visual cue of their presence. Audio can additionally provide semantic information for example, who the person is or what they are saying on the other side of the phone. It can also provide clues about the story, context and sometimes explicitly mention the object or action to complement the video information. Therefore, using audio in video description models will certainly improve the performance~\cite{audioviosionorallba, audiovisualefros}.

\vspace{1mm}
\noindent {\bf External Knowledge:}
In video description, most of the time we are comparing the performance with humans who have extensive out of domain or prior knowledge. When humans watch a clip and describe it, most of the time they don't rely solely on the visual (or even the audio) content. Instead, they additionally employ their background knowledge. Similarly, it would be interesting and promising approach to augment the video description techniques with prior external knowledge~\cite{extknowlangvision}. This approach has shown significantly better performance in visual question answering methods and is likely to improve video description accuracy.

\vspace{1mm}
\noindent {\bf Addressing the Finite Model Capacity:}
Existing methods are trying to perform end-to-end training while using as much data as possible for better learning. However, this approach is inherently limited in learning in itself as no matter how big the training dataset becomes, it will never cover the combinatorial complexity of the real world events. Therefore, learning to use data rather than learning the data itself, is more important and may help improve the upcoming system performances.


\vspace{1mm}
\noindent {\bf Video Description for Subtitle Generation:}
In conjunction with machine translation, video captioning may be used for automatic video subtitling. Currently it is a manual, time consuming, and very costly process. This line of research is not only beneficial for entertainment, one of the largest industries in the world, but it will potentially help improve comprehension of audiovisual material by the visually and hearing impaired, and second language learners.


\vspace{1mm}
\noindent {\bf Automatic Evaluation Measures:}
So far video description has relied on automatic metrics designed for machine translation and image captioning tasks. To date there is no automatic video description (or even captioning) evaluation metric that is purpose designed. Although metrics designed for image captioning are relevant, they have their limitations. This problem is going to exacerbate in the future with dense video captioning and story telling tasks. There is a need for an evaluation metric that is closer to human judgments and that can encapsulate the diversities of realizations of visual content. A promising research direction is to use machine learning to learn such a metric rather than hand engineer it.

\vspace{-2mm}
\section{Conclusion}
\label{sec:conc}
We presented the first comprehensive literature survey of video description research, starting from the classical methods that are based on Subject-Verb-Object (SVO) tuples to more sophisticated statistical and deep learning based methods.  We reviewed popular benchmark datasets that are commonly used for training and testing these models and discussed international competitions/challenges that are regularly held to promote the video description research. We discussed, in detail, the available automatic evaluation metrics for video description, highlighting their attributes and limitations. We presented a comprehensive summary of results obtained by recent methods on the benchmark datasets using all metrics. These results not only show the relative performance of existing methods but also highlight the varying difficulty levels of the datasets and the robustness and trustworthiness of the evaluation metrics. Finally, we put forward some recommendations for future research directions that are likely to push the boundaries of this research area.


From an algorithm design perspective, although LSTMs have shown competitive caption generation performance, the interpretablity and intelligibility of the underlying models are low. Specifically, it is hard to differentiate how much visual features have contributed to the generation of a specific word compared to the bias that comes naturally from the language model adopted. This problem is exacerbated when the aim is to diagnose the generation of erroneous captions. For example, when we see a caption ``red fire hydrant'' generated by a video description model from a frame containing a ``white fire hydrant'', it is difficult to ascertain whether the color feature is incorrectly encoded by the visual feature extractor or is due to the bias in the used language model towards ``red fire hydrants''. Future research must focus on improving diagnostic mechanisms to pin point the problematic part of the architectures so that it can be improved or replaced. 

Our survey shows that a major bottleneck hindering progress along this line of research is the lack of effective and purposely designed video description evaluation metrics. Current metrics have been adopted either from machine translation or image captioning and fall short in measuring the quality of machine generated video captions and their agreement with human judgments. One way to improve these metrics is to increase the number of reference sentences. We believe that purpose built metrics that are learned from the data itself is the key to advancing video description research.

Some challenges come from the diverse nature of the videos themselves. For instance, multiple activities in a video, where captions represent only some activities, could lead to low video description performance of a model. Similarly, longer duration videos pose further challenges since most action features can only encode short term actions such as trajectory features and C3D features~\cite{c3dgenericfeatures} that are dependent on video segment lengths. Most feature extractors are suitable only for static or smoothly changing images and hence struggle to handle abrupt scene changes. Current methods rather simplify the visual encoding part by representing holistic videos or frames. Attention models may further need to be explored to focus on spatially and temporally significant parts of the video. Similarly, temporal modeling of the visual features itself is quite rudimentary in existing methods. Most methods either use mean pooling which completely discards the temporal information or use the C3D model which can only model 15 frames. Future research should focus on designing better temporal modeling architectures that preferably learn in an end-to-end fashion rather than disentangling the visual description from the temporal model and the temporal modeling from language description.

\vspace{-4mm}
\section*{Acknowledgements}
\vspace{-2mm}
The authors acknowledge Marcus Rohrbach (Facebook AI Research) for his valuable input. The research was supported by ARC Discovery Grant DP160101458 and DP150102405.


\vspace{-20mm}
\begin{IEEEbiography}[{\includegraphics[width=1in,height=1.25in,clip,keepaspectratio]{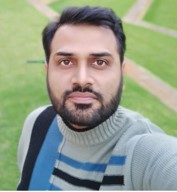}}] 
{Nayyer Aafaq} received BE degree with distinction in Avionics from the College of Aeronautical Engineering (CAE), National University of Sciences and Technology (NUST), Pakistan, in 2007 and MS degree with high distinction in Systems Engineering from Queensland University of Technology (QUT), Australia, in 2012.
He is currently working towards the Ph.D. degree at The University of Western Australia (UWA). He is a recipient of SIRF scholarship at UWA. He has served as a Research Assistant at STG Research Institute, Pakistan, from 2007 to 2011 and as a lecturer at College of Aeronautical Engineering (CAE), NUST, Pakistan from 2013 till 2017. His current research interests includes Deep Learning, Video Analysis and intersection of Natural Language Processing (NLP), Computer Vision (CV) and Machine Learning. 
\end{IEEEbiography}
\vspace{-6mm}
\vskip 0pt plus -1fil
\begin{IEEEbiography}[{\includegraphics[width=1in,height=1.25in,clip,keepaspectratio]{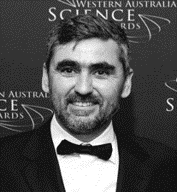}}]
{Ajmal Mian} is a Professor of Computer Science at The University of Western Australia. He completed his PhD from the same institution in 2006 with distinction and received the Australasian Distinguished Doctoral Dissertation Award from Computing Research and Education Association of Australasia. He received the prestigious Australian Postdoctoral and Australian Research Fellowships in 2008 and 2011 respectively. He received the UWA Outstanding Young Investigator Award 2011, the West Australian Early Career Scientist of the Year 2012 award, the Vice-Chancellors Mid-Career Research Award 2014, the Aspire Professional Development Award 2016 and the Excellence in Research Supervision Award 2017. He has published over 160 scientific papers in reputable journals and conferences. He has secured eight Australian Research Council grants, a National Health and Medical Research Council grant and a DAAD German Australian research cooperation grant. He has served as a guest editor of Pattern Recognition, Computer Vision and Image Understanding and Image and Vision Computing journals. His research interests include computer vision, machine learning, 3D shape analysis, hyperspectral image analysis, pattern recognition, and multimodal biometrics.
\end{IEEEbiography}
\vskip 0pt plus -1fil
\begin{IEEEbiography}[{\includegraphics[width=1in,height=1.25in,clip,keepaspectratio]{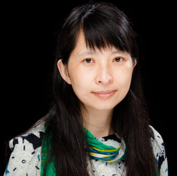}}]
{Wei Liu} received her PhD from the University of Newcastle, Australia in 2003. She is now working at the Department of Computer Science and Software Engineering at the University of Western Australia, and co-lead the faculty's Big Data research group. Her research impact in the field of knowledge discovery from natural language text data is evident by a series of highly cited papers, and the reputable top data mining and knowledge management journals and conferences that she has been published in. These include for example, ACM Computer Surveys, Journal of Data Mining and Knowledge Discovery, Knowledge and Information Systems, International Conference on Data Engineering (ICDE), ACM International Conference on Information and Knowledge Management (CIKM). She has won three Australian Research Council Grants and several industry grants. Her current research focuses on deep learning methods for knowledge graph construction from natural language text, sequential data mining and text mining. 
\end{IEEEbiography}
\vskip 0pt plus -1fil
\vspace{-3mm}
\begin{IEEEbiography}[{\includegraphics[width=1in,height=1.25in,clip,keepaspectratio]{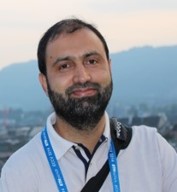}}]
{Syed Zulqarnain Gilani} received his PhD from the University of Western Australia where he is now working as a Research Fellow. He did his MS in EE from the National University of Sciences and Technology (NUST), Pakistan in 2009 and secured the Presidents Gold Medal. His research interests include 3D facial morphometrics with applications to syndrome delineation  and machine learning.
\end{IEEEbiography}
\vskip 0pt plus -1fil
\vspace{-3mm}
\begin{IEEEbiography}[{\includegraphics[width=1in,height=1.25in,clip,keepaspectratio]{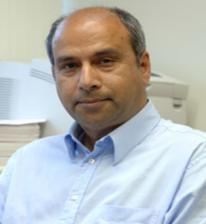}}]
{Mubarak Shah}, the Trustee chair professor of computer science, is the founding
director of the Center for Research in Computer Vision at the University of Central Florida (UCF).
He is an editor of an international book series on video computing, was editor-in-chief of Machine
Vision and Applications journal, and an associate editor of ACM Computing Surveys journal. He was the program cochair of CVPR 2008, an associate editor of the IEEE T-PAMI, and a guest editor of the special issue of the International Journal of Computer Vision on Video Computing. His research interests include video surveillance, visual tracking, human activity recognition, visual analysis of crowded scenes, video registration, UAV video analysis, and so on. He is an ACM distinguished speaker. He was an IEEE distinguished visitor speaker for 1997-2000 and received the IEEE Outstanding Engineering Educator Award in 1997. In 2006, he was awarded a Pegasus Professor Award, the highest award at UCF. He received the Harris Corporations Engineering Achievement Award in 1999, TOKTEN awards from UNDP in 1995, 1997, and 2000, Teaching Incentive Program Award in 1995 and 2003, Research Incentive Award in 2003 and 2009, Millionaires Club Awards in 2005 and 2006, University Distinguished Researcher Award in 2007, Honorable mention for the ICCV 2005 Where Am I? Challenge Problem, and was nominated for the Best Paper Award at the ACM Multimedia Conference in 2005. He is a fellow of the IEEE, AAAS, IAPR, and SPIE. 

\end{IEEEbiography}

\end{document}